\newcommand{\PaperNumber}{475}

\documentclass[10pt,sigconf,letterpaper,nonacm]{acmart}

\usepackage{booktabs}
\usepackage{tabularx}
\usepackage{makecell}
\usepackage{adjustbox}
\usepackage{lscape}
\usepackage{tcolorbox}
\usepackage{graphicx}  
\usepackage{float}
\usepackage[super]{nth}
\usepackage{cleveref}
\usepackage{pifont}
\usepackage{CJKutf8}
\usepackage{subcaption}
\usepackage{hyperref}
\usepackage[unicode]{hyperref}
\hypersetup{
    colorlinks=true,
    linkcolor=blue,
    urlcolor=cyan
}
\newcolumntype{L}{>{\raggedright\arraybackslash}X}
\newcommand{\myparagraph}[1]{\vspace{0.1cm}\textbf{#1}}
\AtBeginDocument{%
  }

\crefformat{section}{\S#2#1#3}
\crefname{figure}{Figure}{Figures}
\crefname{appendix}{Appendix}{Appendices}
\crefname{table}{Table}{Tables}
\crefname{algorithm}{Algorithm}{Algorithms}

\definecolor{green}{HTML}{44AA99}
\definecolor{yellow}{HTML}{DDCC77}
\definecolor{blue}{HTML}{88CCEE}
\definecolor{darkblue}{HTML}{0000FF}
\definecolor{red}{HTML}{CC6677}
\definecolor{darkred}{HTML}{DC3220}
\newcommand{\numberofcountry}{{12}\xspace}
\newcommand{\numberofwebsites}{{120,000}\xspace}

\definecolor{MyForestGreen}{RGB}{34,139,34}
\definecolor{MyMaroon}{RGB}{128,0,0}
\newcommand{\cmark}{{\color{MyForestGreen}\ding{51}}}%
\newcommand{\xmark}{{\color{MyMaroon}\ding{55}}}%

\usepackage{xspace}
\newcommand{\tool}{{\texttt{LangCrUX}}\xspace}
\newcommand{\testingtool}{{\texttt{Kizuki}}\xspace}

\newcommand{\eg}{{e.g.,}\xspace}

\def\*#1*#2{o\null{#2}{#1}}

\def\sh#1{\setbox0=\hbox{#1}%
     \kern-.02em\copy0\kern-\wd0
     \kern.04em\copy0\kern-\wd0
     \kern-.02em\raise.0433em\box0 }
\begin{document}
\pagestyle{plain}
\title{Not All Visitors are Bilingual: A Measurement Study of the Multilingual Web from an Accessibility Perspective}

\author{Masudul Hasan Masud Bhuiyan}
\email{masudul.bhuiyan@cispa.de}
\orcid{0000-0002-7090-4334}
\affiliation{%
  \institution{CISPA Helmholtz Center for Information Security}
  \city{Saarbrücken}
  \country{Germany}
}
 
\author{Matteo Varvello}
\email{matteo.varvello@nokia.com}
\orcid{0000-0001-8500-4630}
\affiliation{%
  \institution{Nokia Bell Labs}
  \city{Holmdel}
  \state{NJ}
  \country{USA}}

\author{Yasir Zaki}
\email{yasir.zaki@nyu.edu}
\orcid{0000-0001-8078-6944}
\affiliation{%
  \institution{New York University Abu Dhabi}
  \city{Abu Dhabi}
  \country{UAE}}

\author{Cristian-Alexandru Staicu}
\email{staicu@cispa.de}
\orcid{0000-0002-6542-2226}
\affiliation{%
  \institution{CISPA Helmholtz Center for Information Security}
  \city{Saarbrücken}
  \country{Germany}
}

\begin{abstract}
English is the predominant language on the web, powering nearly half of the world's top ten million websites. Support for multilingual content is nevertheless growing, with many websites increasingly combining English with regional or native languages in both visible content and hidden metadata. This multilingualism introduces significant barriers for users with visual impairments, as assistive technologies like screen readers frequently lack robust support for non-Latin scripts and misrender or mispronounce non-English text, compounding accessibility challenges across diverse linguistic contexts. Yet, large-scale studies of this issue have been limited by the lack of comprehensive datasets on multilingual web content. To address this gap, we introduce \tool, the first large-scale dataset of \numberofwebsites popular websites across \numberofcountry languages that primarily use non-Latin scripts. Leveraging this dataset, we conduct a systematic analysis of multilingual web accessibility and uncover widespread neglect of accessibility hints. We find that these hints often fail to reflect the language diversity of visible content, reducing the effectiveness of screen readers and limiting web accessibility. We finally propose \testingtool, a language-aware automated accessibility testing extension to account for the limited utility of language-inconsistent accessibility hints.
\end{abstract}

\maketitle
%%%%%%%%%%%%%%%%%%%%%%%%
\section{Introduction}
%%%%%%%%%%%%%%%%%%%%%%%%

An estimated 2.2 billion people live with some form of visual impairment, and 90\% of them reside in low- and middle-income countries~\cite{who2021}. As internet adoption grows in these regions, an increasing share of web content is created in languages other than English. Many of these languages use non-Latin writing systems such as Devanagari, Bengali, Arabic, or Thai and are often presented alongside English in mixed-language interfaces~\cite{pimienta2022resource, pimienta2023method, pimientareliably}. Popular screen readers like JAWS~\cite{freedomscientificJAWSxAEx2013} and NVDA~\cite{nvaccessAccess} still exhibit limited support for non-Latin scripts and often perform poorly when confronted with mixed languages~\cite{auer2013code,mccarthy_analysis_2012, vashistha_educational_2014, vashistha2016technology}.  
These tools may misrender non-English words or produce unintelligible output, as has been documented with languages like Nepali~\cite{sankhi_glimpse_2022}.

The root of the problem lies in the inadequate usage of \textit{text alternatives}~\cite{digitala11yUnderstandingWCAG, w3ContentAccessibility} metadata such as \texttt{lang} attributes or image \texttt{alt text}, which screen readers use to process content appropriately. When such metadata is absent, incorrect, or inconsistent with the visible text, it creates a mismatch between the displayed content and what the assistive tool offers. This issue is compounded for scripts that require complex shaping or language-specific pronunciation models. For example, Apple's VoiceOver~\cite{appleVoiceOverUser}, the default screen reader on macOS and iOS devices, does not provide any support for languages such as Urdu, Amharic, or Burmese~\cite{appleVoiceOverUser}. This lack of linguistic inclusivity is at odds with the principles laid out by the W3C Web Accessibility Initiative, which advocates for equal digital access regardless of language or ability~\cite{wcag2023}.

Web accessibility research only marginally studies the intersection of multilingualism and assistive technology~\cite{garcia2024multilingual,uxdesignTroubledState}. A key bottleneck in this area is the lack of data. Widely used datasets like  Tranco~\cite{pochat2018tranco} focus on popularity but do not provide any insight into the linguistic composition of websites. This limits the researchers' ability to measure the prevalence of multilingual content and to assess the  accessibility landscape for speakers of less commonly supported languages. 
To bridge this gap, we introduce \tool, a large-scale dataset of \numberofwebsites popular websites across \numberofcountry languages that primarily use non-Latin scripts. We construct LangCrUX by selecting high-traffic websites from the Chrome User Experience Report (CrUX) dataset, verifying language use via automated script detection and manual sampling, and through Puppeteer-based crawls routed via country-specific VPNs to capture localized versions of each site.

We leverage \tool to conduct the first large-scale analysis of multilingual web accessibility, focusing on how assistive technologies interact with real-world, multilingual web content. Our findings reveal that nearly 40\% of websites in Bangladesh and India lack any accessibility text in the native language, despite having predominantly native-language visible content. More broadly, language-aware accessibility remains a widespread and under-addressed issue: many websites use non-descriptive, placeholder, or untranslated text (\eg ``file1,'' ``button,'' or generic English terms) in critical accessibility elements like image \texttt{alt text}. 
These mismatches significantly reduce the utility of screen readers, which rely on  metadata to convey meaningful information to users.
We observe that current automated accessibility testing tools fail to detect these language inconsistencies, as they typically only evaluate the presence of accessibility hints. To address this gap, we propose and develop \testingtool\footnote{named after the Japanese word for ``awareness''}, a testing extension that identifies such mismatches and evaluates metadata based on alignment with the surrounding linguistic context, offering a more inclusive measure of accessibility.

%%%%%%%%%%%%%%%%%%%%%%%%%%
\section{Methodology}
\label{sec:meth}
\begin{figure}
    \centering
    \includegraphics[width=\linewidth]{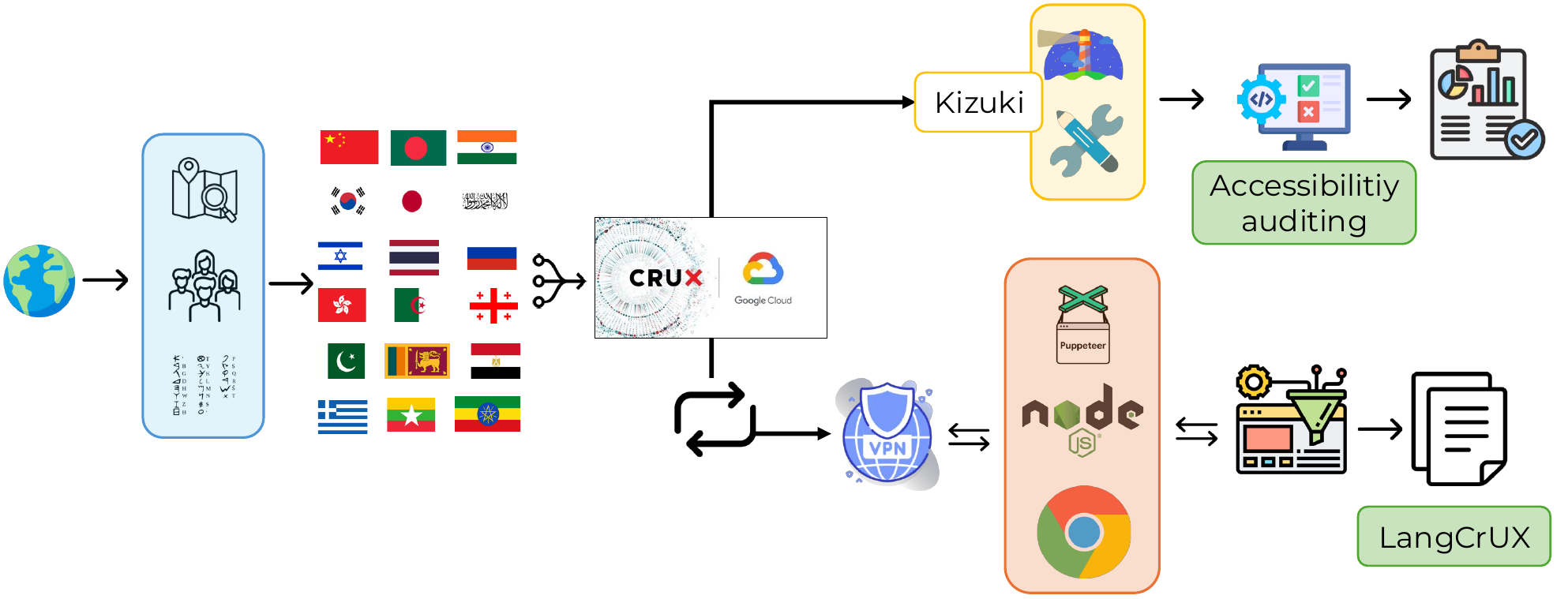}
    \vspace{-5mm}
    \caption{Overview of our methodology for constructing and analyzing the \tool dataset.}
    \label{fig:methodology}
    \vspace{-5mm}
\end{figure}

%%%%%%%%%%%%%%%%%%%%%%%%%%
Our study investigates the accessibility of websites in non-Latin-script languages, focusing on linguistic diversity in underrepresented language communities. We collect \textbf{\tool}, a dataset comprising websites with a high proportion of non-Latin-script content, combining data from the Chrome User Experience Report (CrUX)~\cite{crux}, custom web crawls using Puppeteer, and language metadata verification. Figure~\ref{fig:methodology} visualizes our methodology, including dataset construction, language selection, and automated accessibility analysis.
 \tool is released as an open-source dataset on GitHub.\footnote{\url{https://anonymous.4open.science/r/LangCrux-F68F}}

%%%%%%%%%%%%%%%%%%%%%%%%%%%%%%%%%%%%%%%%%%%%%%%%%%%%
\vspace{5pt}
\noindent\textbf{Language and Country Selection Criteria:} 
%%%%%%%%%%%%%%%%%%%%%%%%%%%%%%%%%%%%%%%%%%%%%%%%%%%%
We begin with a pool of 26 widely spoken non-Latin-script languages, including Hindi, Bangla, Modern Standard Arabic, Tamil, Telugu, Mandarin Chinese, Urdu, Amharic, Russian, Marathi, and others~\cite{wikipediaListLanguages}. Our goal is to identify a diverse, representative set of languages underrepresented in web accessibility research, particularly those using non-Latin writing systems. Language selection is guided by three main factors: script type (non-Latin), size of the global speaker base, and geographic and linguistic diversity. To ensure representativeness and sufficient data, we apply two strict inclusion criteria: 1) at least 10{,}000 websites with 50\% or more visible textual content in the target language, and 2) inclusion in the CrUX dataset -- which provides user experience metrics from Chrome users -- with sufficient traffic and performance data. Applying these filters results in a final set of \numberofcountry language-country pairs, each uniquely representing a distinct non-Latin-script language with a verifiable web presence. 
Because the initial pool alone did not yield enough languages that met these thresholds, we expanded the selection to include additional ones such as Hebrew, Sinhala, Greek, and Burmese. These were added to increase script and regional diversity while still satisfying our inclusion criteria.

\begin{figure}
    \centering
    \begin{subfigure}[b]{0.48\linewidth}
        \includegraphics[width=\linewidth]{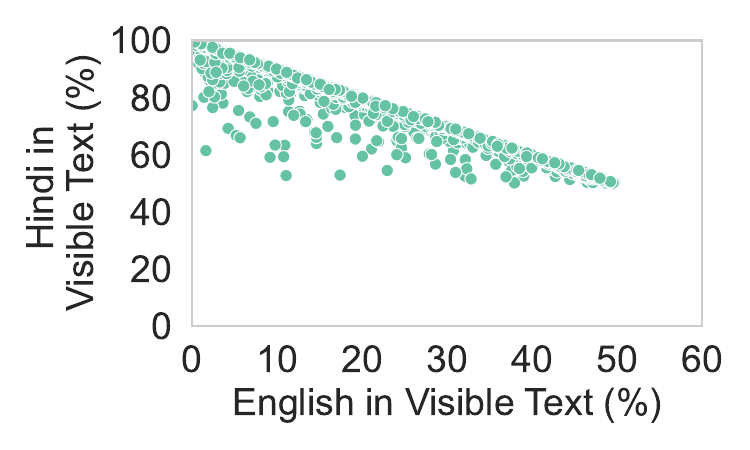}
        \caption{Hindi (India)}
        \label{fig:visible_text}
    \end{subfigure}
    \hfill
    \begin{subfigure}[b]{0.48\linewidth}
        \includegraphics[width=\linewidth]{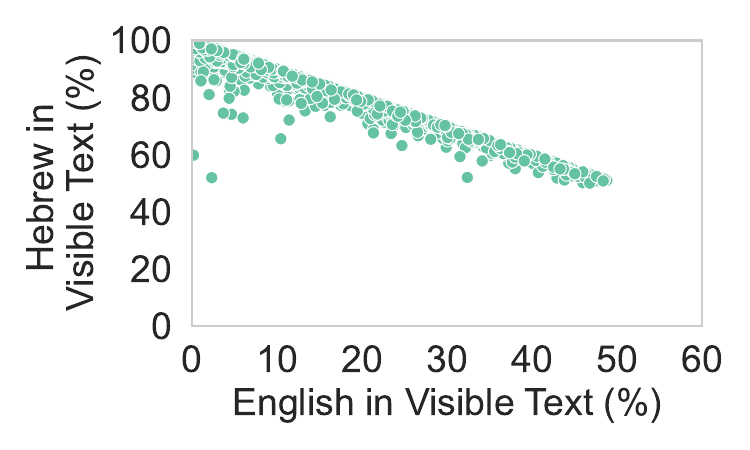}
        \caption{Hebrew (Israel)}
        \label{fig:accessible_text}
    \end{subfigure}
    \vspace{-3mm}
    \caption{Native language distribution in visible text for India and Israel in \tool}
    \label{fig:lang_distribution}
    \vspace{-5mm}
\end{figure}

The selected countries, their corresponding languages, and approximate global speaker populations are China (Mandarin Chinese, 1.2 billion), India (Hindi, 609 million), Algeria (Modern Standard Arabic, 335 million), Bangladesh (Bangla, 284 million), Russia (Russian, 253 million), Japan (Japanese, 126 million), Egypt (Egyptian Arabic, 119 million), Hong Kong (Cantonese, 85.5 million), South Korea (Korean, 82 million), Thailand (Thai, 71 million), Greece (Greek, 13.5 million), and Israel (Hebrew, 9 million). Collectively, these \numberofcountry languages are spoken by over 3.19 billion people, representing about 39.5\% of the global population. For languages spoken in multiple countries, such as Modern Standard Arabic, used in Algeria, Saudi Arabia, and Morocco, we select the country with the highest population of native speakers, in this case, Algeria.
In multilingual countries like India, we include all major non-Latin-script languages with substantial speaker populations. However, only Hindi meets our data threshold; other widely spoken languages, such as Tamil and Telugu, do not meet the 10{,}000-website requirement and are excluded. Similar exclusions apply to Sinhala (Sri Lanka) and Georgian (Georgia), where websites with sufficient native-language content fell below the threshold despite initial inclusion.

%%%%%%%%%%%%%%%%%%%%%%%%%%%%%%%%%%%%%%%%%%%%%%%%%%%%
\vspace{5pt}
\noindent\textbf{Website Selection:} 
%%%%%%%%%%%%%%%%%%%%%%%%%%%%%%%%%%%%%%%%%%%%%%%%%%%%
We use Google CrUX to identify and rank websites by real-world usage metrics. For each selected language-country pair, we extract the top 10{,}000 websites based on CrUX rankings, which reflect user engagement, load performance, and interaction quality. 
To validate language presence, we use a Unicode-based heuristic that matches visible text content against script-specific character ranges (e.g., Devanagari for Hindi, Hangul for Korean, and Cyrillic for Russian). For overlapping scripts, such as Arabic and Urdu, we include additional language-specific characters to improve precision.
A website is retained if at least 50\% of its visible textual content is in the target language. Websites that do not meet this threshold are excluded and replaced with the next-ranking candidate from the CrUX list. In cases where 10{,}000 qualifying websites could not be found among the top-ranked entries, we extended our search to lower-ranked websites within the CrUX database to fulfill the quota. Figure~\ref{fig:lang_distribution} illustrates the distribution of visible content by language for two representative cases: Hindi websites in India and Hebrew websites in Israel.

%%%%%%%%%%%%%%%%%%%%%%%%%%%%%%%%
\vspace{5pt}
\noindent\textbf{Data Collection:} 
%%%%%%%%%%%%%%%%%%%%%%%%%%%%%%%%
To extract accessibility-related features from the selected websites, we develop a web crawler using Puppeteer~\cite{pptrPuppeteerPuppeteer}, which simulates web browsing conditions in a Chromium environment. Each website is visited programmatically, allowing us to capture network-level metadata, page structure, and accessibility indicators such as alternative text, ARIA (Accessible Rich Internet Applications) attributes (which enhance the semantics of web elements for assistive technologies~\cite{mozillaARIAAccessibility}), and declared language tags.

To capture the localized experience of users in each country, we route all browser traffic through VPN servers physically hosted in the corresponding country. This step is critical for collecting region-specific versions of websites, as many sites dynamically serve content---including language settings, layout, or accessibility features---based on the user's IP location. Without VPN-based localization, web crawlers risk accessing global or English-dominant versions of websites that do not accurately reflect the intended user experience of native speakers.
We use a combination of commercial VPN services, including ProtonVPN~\cite{protonvpnBestSpeed} and Hotspot Shield~\cite{hotspotshieldHotspotShield}, to achieve broad geographic coverage. Since not all VPN providers have servers in every target country, we select the provider on a per-country basis to ensure reliable and consistent access from within national borders. Compared to crawling from generic cloud-hosted IPs, this approach offers a significantly more realistic vantage point, reducing the risk of content variation, redirection, or censorship artifacts that may otherwise skew accessibility analysis.

%%%%%%%%%%%%%%%%%%%%%%%%%%%%%%%%%%%%%%%%%%%%%%%%%%%%%%%%%%%%%%%%
\vspace{5pt}
\noindent\textbf{Accessibility Element Selection Criteria:}
\label{subsec:element_selection}
%%%%%%%%%%%%%%%%%%%%%%%%%%%%%%%%%%%%%%%%%%%%%%%%%%%%%%%%%%%%%%%%
To identify accessibility elements where natural language plays a critical role, we follow a structured process based on the Lighthouse accessibility auditing framework~\cite{lighthouse}. Lighthouse evaluates a range of accessibility checks, each associated with specific HTML elements and best practices. Our goal is to select those elements for which the presence, clarity, and appropriateness of natural language directly influence accessibility outcomes.

We begin by examining the set of Lighthouse accessibility tests, which internally relies on the Axe-core accessibility engine~\cite{dequeuniversityListRules}. For each test, we identify the corresponding HTML element it targets (\eg \texttt{<img>}, \texttt{<button>}, \texttt{<input>}). We then analyze the test rationale by referencing the associated rule definitions from Axe-core, which provide detailed explanations and criteria for each audit. From these specifications, we assess whether natural language content is integral to the test. Specifically, we ask whether the accessibility of the element depends on human-readable text, for example, whether a screen reader user would rely on the clarity and relevance of that text to understand the element's purpose. If natural language is central to the function or evaluation of the element, we include it in our set. Following this process, we identify the twelve elements listed in Table~\ref{tab:language_sensitive_elements} as language-sensitive accessibility features.
\begin{table}
    \centering
    \begin{tabular}{lll}
    \toprule
     \texttt{button-name} & \texttt{document-title} & \texttt{image-alt} \\
     \texttt{frame-title} & \texttt{summary-name} & \texttt{label}\\
     \texttt{input-image-alt} & \texttt{select-name} & \texttt{link-name}\\
     \texttt{input-button-name} & \texttt{svg-img-alt} & \texttt{object-alt} \\
    \bottomrule
    \end{tabular}
    \caption{Web elements requiring natural language. } 
    \vspace{-10mm}
    \label{tab:language_sensitive_elements}
\end{table}

These elements span a range of interface components, including images, forms, buttons, and navigation elements, all of which rely on meaningful textual descriptions to be accessible to users with visual impairments. We explicitly exclude certain tests, such as \texttt{video-caption}. Although captions are inherently language-dependent and critical for accessibility, accurately evaluating them at scale poses challenges. In many cases, captions are not embedded in the HTML but are provided through separate files (e.g., VTT or SRT) or dynamically loaded via JavaScript. These may be inaccessible to crawlers unless the video is played or fully rendered in the browser. Furthermore, identifying whether a video has accurate, synchronized, and complete captions often requires manual inspection or audio-visual comparison—steps outside the scope of automated large-scale analysis. Due to these limitations, we omit video-caption checks to maintain consistency and reproducibility in our methodology.

%%%%%%%%%%%%%%%%%%%%%%%%%%%%
\vspace{5pt}
\noindent\textbf{Limitations:} 
%%%%%%%%%%%%%%%%%%%%%%%%%%%%
Our methodology has several limitations. First, reliance on CrUX limits our scope to websites with measurable Chrome traffic, potentially excluding low-traffic or highly localized websites. Next, while using a VPN allows accessing region-specific versions of websites, some websites may detect VPN use and return generic or restricted versions. In such cases, we replace the affected websites with the next eligible candidate. Additionally, Puppeteer's simulated browsing environment may not fully reflect user experiences. Finally, language detection (both automated and manual) can be challenged by embedded content or non-standard scripts, though our 50\% content threshold and manual verification aim to reduce this risk.

%%%%%%%%%%%%%%%%%%%%%%%%%%%%%%%%%%%%%%%%%%%%%%%
\section{Is Multilingual Web Accessible?}
\label{sec:res}
%%%%%%%%%%%%%%%%%%%%%%%%%%%%%%%%%%%%%%%%%%%%%%%

\begin{table*}
    \centering
    \begin{adjustbox}{max width=\textwidth, center}
    \begin{tabular}{lrrrrrrrrrrrr}
        \toprule
        Element & \multicolumn{3}{c}{Missing (\%)} & \multicolumn{3}{c}{Empty (\%)} & \multicolumn{3}{c}{Text Length} & \multicolumn{3}{c}{Word Count} \\
        \cmidrule(r){2-4} \cmidrule(r){5-7} \cmidrule(r){8-10} \cmidrule(r){11-13}
         & Median & Std Dev & Mean & Median & Std Dev & Mean & Median & Std Dev & Mean & Median & Std Dev & Mean \\
        \midrule
        button-name       & 71.43 & 37.25 & 61.92 & 0.00 & 4.18 & 0.36 & 14 & 25.01 & 21.35 & 3 & 4.72 & 3.83 \\
        frame-title       & 87.50 & 30.09 & 75.81 & 0.00 & 3.33 & 0.21 & 13 & 11.57 & 17.45 & 1 & 2.39 & 2.54 \\
        image-alt         & 1.89 & 28.86 & 17.12 & 7.46 & 32.40 & 25.39 & 14 & 1332.15 & 22.97 & 2 & 8.27 & 3.67 \\
        input-button-name & 100.00 & 22.62 & 93.90 & 0.00 & 4.03 & 0.19 & 12 & 13.12 & 14.26 & 2 & 2.70 & 2.83 \\
        input-image-alt   & 0.00 & 47.17 & 35.07 & 0.00 & 21.27 & 4.85 & 3 & 6.92 & 5.66 & 1 & 1.26 & 1.41 \\
        label             & 100.00 & 10.01 & 98.55 & 0.00 & 1.27 & 0.02 & 8 & 6.49 & 9.28 & 1 & 1.04 & 1.67 \\
        link-name         & 100.00 & 11.98 & 95.96 & 0.00 & 0.98 & 0.04 & 22 & 27.01 & 26.64 & 3 & 4.39 & 4.67 \\
        object-alt        & 100.00 & 23.30 & 94.19 & 0.00 & 5.09 & 0.26 & 9 & 17.64 & 14.26 & 1 & 3.48 & 2.49 \\
        select-name       & 100.00 & 28.78 & 89.84 & 0.00 & 2.00 & 0.05 & 10 & 23.03 & 12.94 & 2 & 2.26 & 2.30 \\
        summary-name      & 100.00 & 25.84 & 90.47 & 0.00 & 2.98 & 0.17 & 5 & 3.68 & 5.69 & 1 & 0.59 & 1.18 \\
        svg-img-alt       & 100.00 & 15.15 & 96.66 & 0.00 & 2.90 & 0.15 & 13 & 5.38 & 11.98 & 2 & 0.74 & 1.88 \\
        \bottomrule
    \end{tabular}
    \end{adjustbox}
    \caption{Accessibility element statistics showing \textbf{median}, \textbf{standard deviation}, and \textbf{mean} values for missing and empty percentages, as well as descriptive richness (text length and word count).}
    \label{tab:overview_table}
    \vspace{-0.2in}
\end{table*}

Table~\ref{tab:overview_table} provides statistics on the quality and presence of accessibility text across twelve HTML elements. For each element, we report the median, standard deviation, and average percentage of websites where the accessibility attribute is missing or empty. We also include metrics like text length (in characters) and word count to assess richness and verbosity. These measurements allow us to compare how frequently accessibility features are implemented and how informative they are when present.
In the following, we analyze these patterns by examining missing and empty values, evaluating text length and word count, filtering uninformative content, and characterizing the language distribution of informative accessibility text.

\myparagraph{Prevalence of Missing and Empty Accessibility Texts:}
Missing accessibility text is a widespread issue across HTML elements. Several elements exhibit extremely 
high average missing rates, including \texttt{label} (98.5\%), \texttt{svg-img-alt} (96.6\%), \texttt{link-name} (95.9\%), \texttt{input-button-name} (93.9\%), and \\ \texttt{summary-name} (90.47\%). 
The \texttt{image-alt} attribute stands out for its relatively lower average missing rate (17.12\%) but exhibits the highest percentage of empty values (25.39\%). Although our findings show that it is possible to pass the Lighthouse audit for image accessibility by setting the \texttt{alt} attribute to an empty string, such text does not convey meaningful information to users (\cref{subsec:appendix-lighthouseevalution}). Compared to non-multilingual websites, the missing percentage is slightly higher (15.19\% vs. 17.12\%) and the empty percentage is notably higher (16.36\% vs. 25.39\%), indicating a greater tendency to include but not meaningfully populate the attribute in multilingual contexts.

Attributes like \texttt{button-name} and \texttt{link-name}, which are essential for identifying interactive components, also show high missing rates (61.92\% and 95.96\%, respectively). Similarly, attributes associated with form controls, such as \texttt{input\allowbreak -\allowbreak button\allowbreak -\allowbreak name} and \texttt{select-name}, are frequently absent or left empty, compromising the clarity of form interactions.
A likely reason for these high rates is that, in many cases, screen readers fall back to reading visible HTML text (such as the inner text of a button or link) when accessibility attributes like \texttt{aria-label}, \texttt{label}, or \texttt{alt} are missing. 
This fallback behavior reduces the perceived need for developers to explicitly include accessibility metadata, especially when the element already includes visible text. 

\myparagraph{Text Length and Word Count Analysis:} Table~\ref{tab:overview_table} also captures the descriptive quality of accessibility text through text length and word count metrics. Among the elements, \texttt{link-name} has a relatively high average text length ($\sim$27 characters) and word count ($\sim$5), compared to \texttt{summary-name}, \texttt{select-name}, \texttt{label}, and \texttt{button-name}, which show lower average word counts (1.17, 2.31, 1.67, and 3.86, respectively). This suggests that link descriptions tend to be more detailed and contextually informative. However, for some other elements, shorter texts are often acceptable; for example, buttons labeled ``Login,'' ``Send,'' or ``Submit'' typically provide sufficient clarity with just one or two words.

For elements like \texttt{image-alt}, which require contextual descriptions to convey the meaning of an image, we find an average word count of only $\sim$4. This shortness, especially when combined with the high empty rate noted earlier, suggests a broader trend of developers including minimal or superficial alt text, possibly to satisfy automated checks rather than to genuinely enhance accessibility. Finally, the table reveals substantial outliers, \eg \texttt{image-alt} has a maximum of 261,864 characters and 12,306 words, while \texttt{link-name} 5{,}228 characters and 518 words. These extreme but rare cases likely indicate instances where extraneous content, such as metadata, boilerplate text, or full paragraphs, has been mistakenly inserted into accessibility attributes, potentially overwhelming assistive technologies and undermining user experience (see ~\cref{appendix:outlier-examples} for examples). 

\begin{figure}
    \centering
    \includegraphics[width=\linewidth]{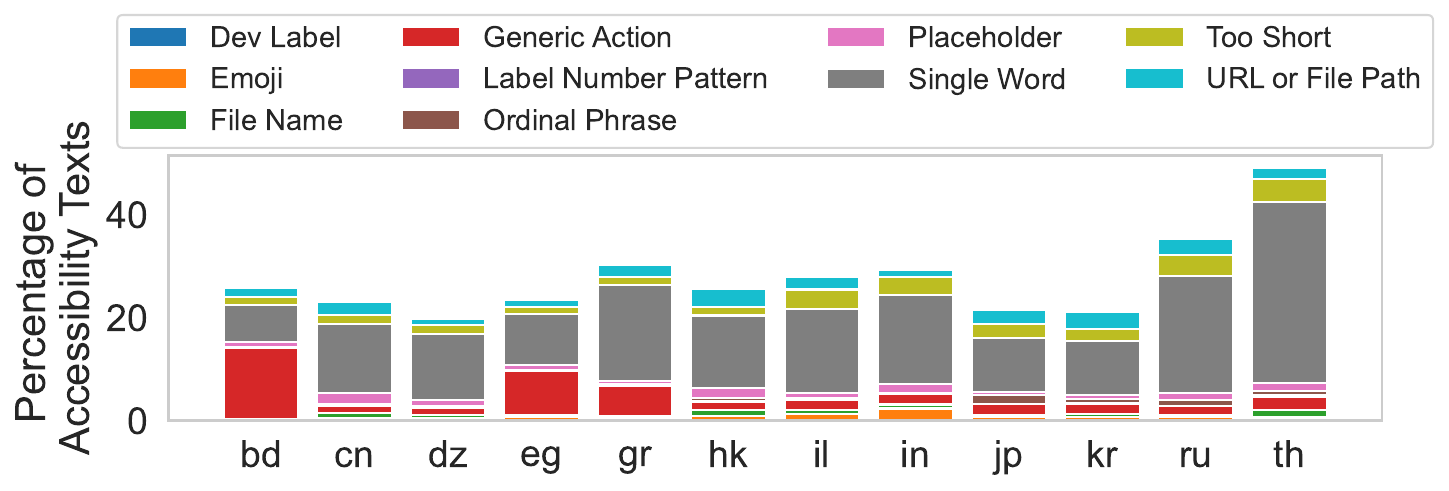}
    \vspace{-7mm}
    \caption{Distribution of filtered accessibility texts by discard reason across countries.}
    \label{fig:filtered_reason_percent}
    \vspace{-8mm}
\end{figure}

\vspace{5pt}
\noindent\textbf{Filtering Uninformative Accessibility Text:}  To assess the quality of accessibility text, we apply a filtering step to discard uninformative or placeholder texts. This is essential because the presence of an \texttt{alt} or \texttt{aria-label} attribute does not guarantee usefulness. Labels such as \texttt{button}, \texttt{file1}, or \texttt{image1} may satisfy automated checks but provide no semantic value to screen reader users. We define a set of heuristics to classify accessibility texts into eleven categories, distinguishing between useful and discardable content. These include short strings, file paths, placeholders, developer labels, etc. \cref{appendix:filtering} shows the full list of filtering rules. ~\cref{appendix:element_filtering} shows the breakdown by HTML element.

Figure~\ref{fig:filtered_reason_percent} shows the percentage distribution of filtered accessibility texts across countries and categories. One of the most common issues is the use of generic single words. For example, in Thailand, over 33\% of accessibility texts are single-word labels. High rates are also observed in Russia (22.2\%), Greece (18.03\%), and India (17.1\%). In contrast, countries like Bangladesh (6.9\%) and Egypt (10.5\%) show lower proportions. A small but non-negligible portion of texts are too short to convey meaning. In Russia, 4.26\% of labels fall into this category, followed by Thailand (4.24\%), Israel (4.03\%), and India (3.6\%). Some websites also use raw URLs or file paths as labels. This affects 3.8\% of labels in Hong Kong, 3.5\% in South Korea, and 3.17\% in Russia.
These findings show that even when accessibility text is present, it often lacks descriptive content. This reinforces the need to go beyond presence-based metrics and assess the actual semantic value of accessibility labels.

\begin{figure}
    \centering
    \includegraphics[width=.9\linewidth]{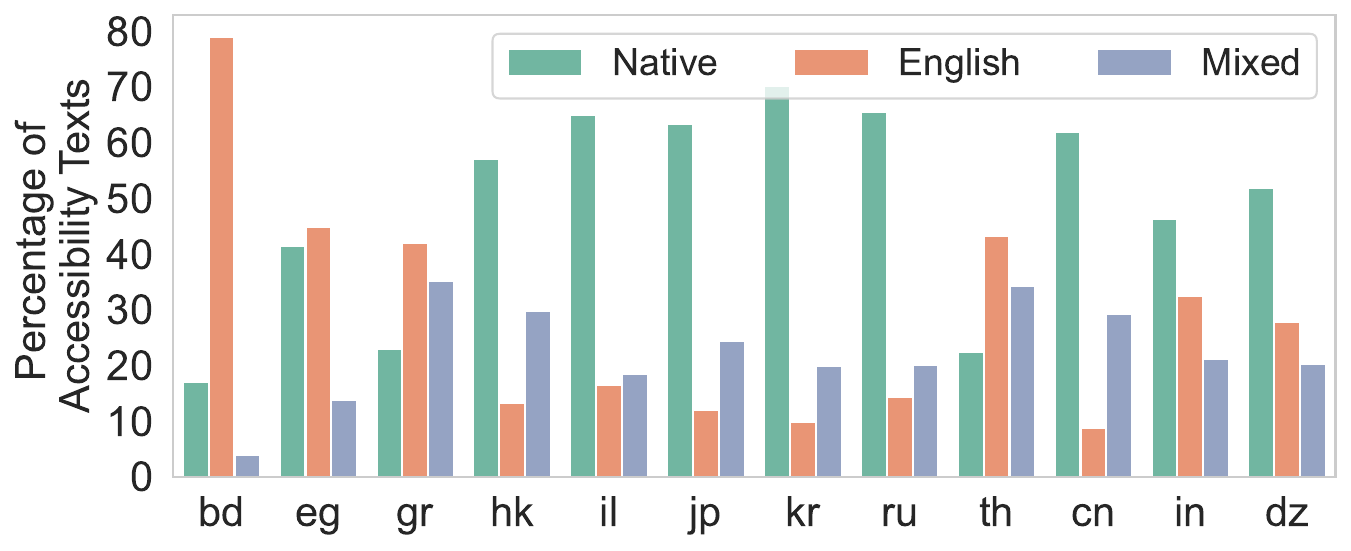}
    \vspace{-3mm}
    \caption{Language distribution of filtered accessibility texts across countries. Only texts classified as potentially useful are included.}
    \label{fig:lang_distribution_filtered}
    \vspace{-7mm}
\end{figure}

\vspace{5pt}
\noindent\textbf{Language Distribution of Informative Accessibility Text:} After removing uninformative and placeholder accessibility text, we reanalyze the remaining content to understand the language distribution of texts that are potentially useful. This filtered set reflects more intentional and meaningful uses of \texttt{alt}, \texttt{label}, and other accessibility attributes. Figure~\ref{fig:lang_distribution_filtered} shows the proportion of accessibility texts written in native languages, English, or a mix of both, across the 12 analyzed countries. 
A prominent pattern is the heavy reliance on English, even in countries where it is not the primary language. In Bangladesh, 79\% of informative accessibility texts are in English, the highest among all countries analyzed. Egypt, Thailand, and Greece also show a strong tendency to default to English. This suggests that developers may rely on English for accessibility metadata due to limited localization tools or being unaware of screen reader needs in native languages.

Another important pattern is the use of mixed-language accessibility hints, where a single \texttt{alt} attribute contains both the native language and English. This occurs frequently in Greece (35\%), Thailand (34\%), and Hong Kong (30\%), and in over 20\% of websites in China, Russia, Japan, and India. While sometimes intended to aid multilingual users, such mixing often confuses screen readers, which typically do not handle language switching within a single label, resulting in mispronunciations or reduced clarity.

%%%%%%%%%%%%%%%%%%%%%%%%%%%%%%%%%%%%%%%%%
\section{Language-Aware Accessibility}
%%%%%%%%%%%%%%%%%%%%%%%%%%%%%%%%%%%%%%%%%
\begin{figure*}
    \centering
    \includegraphics[width=0.95\textwidth]{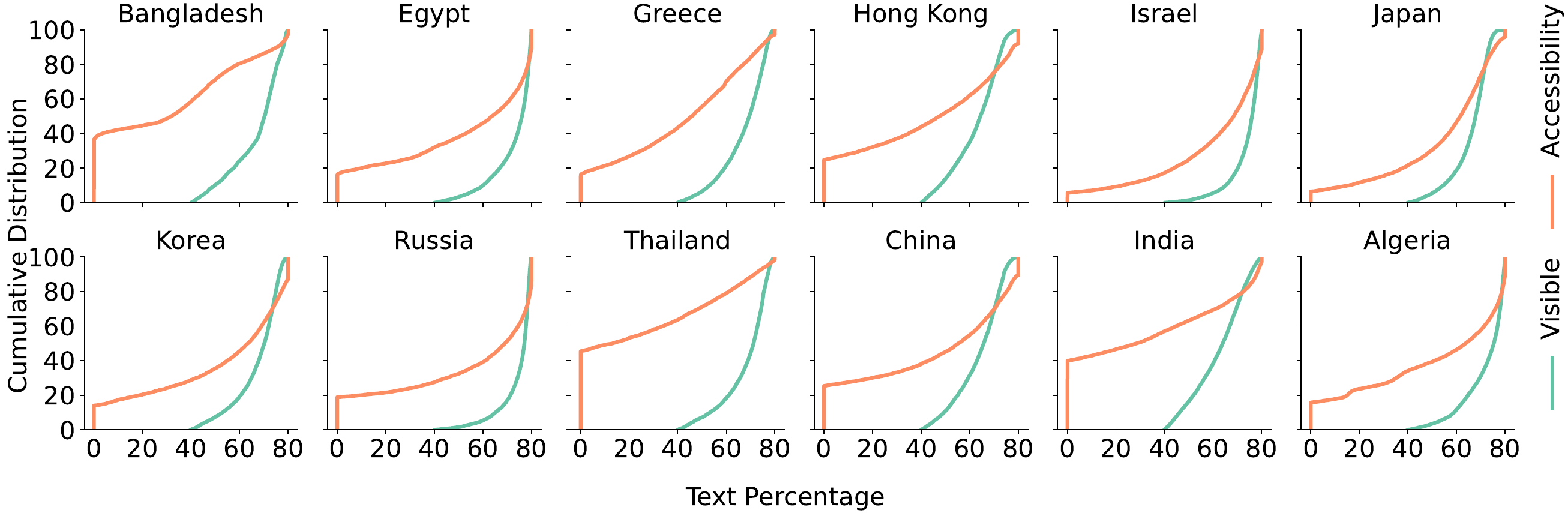}
    \vspace{-3mm}
    \caption{CDFs of native language usage in visible vs. accessibility text across countries. Most websites with visible content in native languages still rely on English in accessibility metadata.}
    \label{fig:text_distribution}
\end{figure*}
\vspace{5pt}

\noindent\textbf{Mismatch Between Visible and Accessibility Text:} Figure~\ref{fig:text_distribution} compares native language usage in visible versus accessibility text across the \numberofcountry analyzed countries. While the visible content of many websites is multilingual or predominantly in the native language, the associated accessibility metadata, such as \texttt{alt} text, \texttt{aria-labels}, and form labels, is typically written in English.  This mismatch is especially noticeable in countries like India and Bangladesh, where over 40\% of websites have less than 10\% of their accessibility text in the native language. Thailand, China, and Hong Kong also show similar trends, with more than a quarter of their websites falling into this category. In contrast, countries like Japan and Israel have significantly lower rates of mismatch, with fewer than 10\% of websites showing such disparities. For blind users who rely on screen readers, this language discrepancy introduces an additional barrier, forcing them to navigate a bilingual interface where visible content and assistive text do not align. ~\cref{appendix:scatterplots} provides a detailed view of the mismatch by visualizing the distribution of visible versus accessibility text across all countries in our dataset.

For example, in Bangladesh, \url{https://teachers.gov.bd} is a widely used government education portal. Although more than 98\% of its visible content is in Bangla, only one of the 79 images with \texttt{alt} attributes uses Bangla. Comparable patterns are also seen in India, Thailand, and China. The Indian website \url{https://cmhelpline.mp.gov.in} has a Hindi version where the interface is almost entirely in Hindi, but all accessibility text is in English. The Thai news website \url{https://www.khaosod.co.th} contains over 92\% of its visible content in Thai, but its accessibility labels are mostly in English. The Chinese provincial government site \url{https://kjt.shaanxi.gov.cn} is almost fully in Chinese, yet its accessibility texts are entirely in English.
Mismatch examples are provided in ~\cref{appendix:mismatch_examples}.

\vspace{5pt}
\noindent\textbf{Adding Language Awareness to Lighthouse:}  Automated  testing tools such as Lighthouse do not consider the language of accessibility text when evaluating compliance. As a result, alt attributes are marked as present regardless of whether their content matches the language of the surrounding interface. To address this limitation, we introduce \textbf{\testingtool}, a Lighthouse extension that incorporates language awareness in accessibility evaluation. Specifically, we extend the audit for image alt text to verify whether the description is written in the same language as the page's visible content.

\begin{figure}
    \centering
    \includegraphics[width=.8\linewidth]{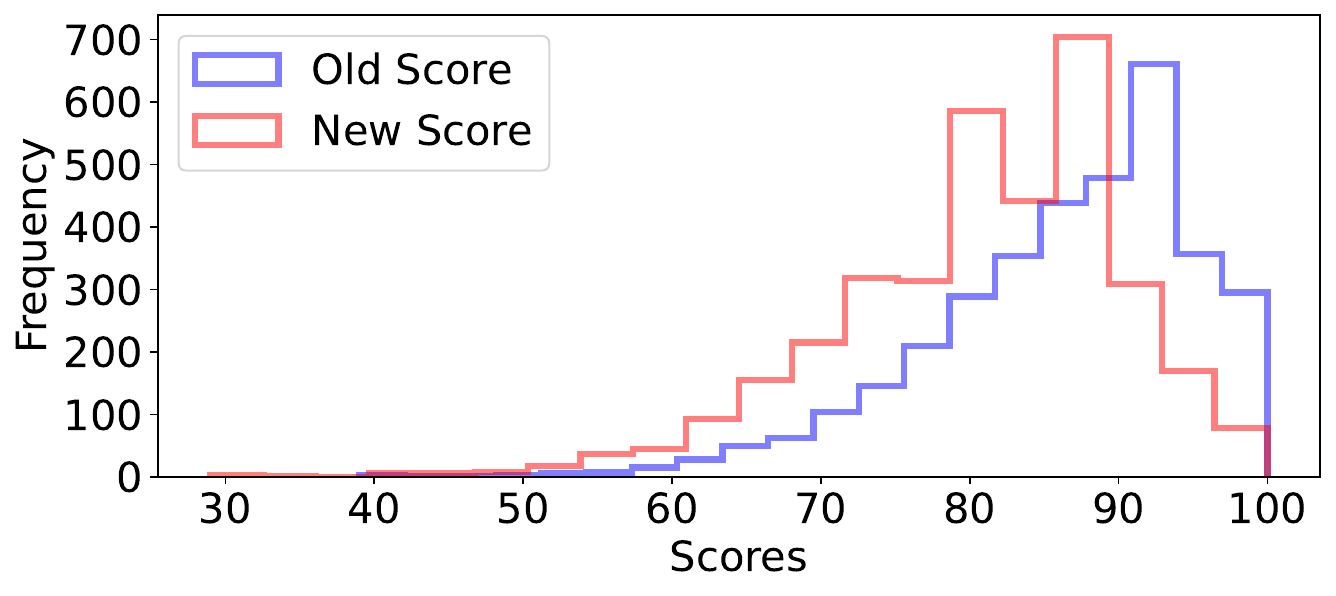}
    \vspace{-5mm}
    \caption{Accessibility score distribution before and after applying Kizuki’s language-aware alt text evaluation on websites from Bangladesh and Thailand.}
    \label{fig:score_distribution}
    \vspace{-5mm}
\end{figure}

We evaluate \testingtool on 10{,}000 websites from Bangladesh and Thailand, two countries where language mismatch between visible content and accessibility metadata is particularly common.
For fairness, we exclude websites that fail the original Lighthouse test due to missing alt attributes. Figure~\ref{fig:score_distribution} shows the resulting shift in accessibility scores. Without considering language, 43\% of websites received a Lighthouse score above 90 (considered ``good''~\cite{chromeLighthousePerformance, graphiteLighthousePerformance}), and 5.6\% achieved a perfect score. After applying Kizuki’s language-aware check, these numbers dropped significantly: only 15.8\% of websites scored above 90, and just 1.8\% retained a perfect score.

%%%%%%%%%%%%%%%%%%%%%%%%
\section{Related Work}
%%%%%%%%%%%%%%%%%%%%%%%%
Several studies have explored the challenges of multilingual web accessibility. Vázquez et al.~\cite{rodriguez_vazquez_introducing_2014, vazquez_unlocking_2015, rodriguez_vazquez_exploring_2015, rodriguez_vazquez_measuring_2016} conducted user studies highlighting the limitations of screen readers in handling multilingual interfaces. 
Casalegno~\cite{casalegno2018usability} reported similar findings, emphasizing the cognitive strain users face when navigating mixed-language content. García-Garcinuño et al.~\cite{garcia2024multilingual} confirmed that language mismatches in screen reader output persist even when markup is correctly annotated. Vázquez et al.~\cite{antona_bringing_2017, rodriguez_vazquez_assuring_2016} also showed that accessibility is often excluded from localization workflows, leading to untranslated alt text and inconsistent metadata. Several works highlight problems such as mispronunciation, broken accents, or robotic voices in the context of multilingual content for screen readers~\cite{mccarthy_analysis_2012, vashistha_educational_2014, vashistha2016technology}. Sankhi et al.~\cite{sankhi_glimpse_2022} documented similar barriers in Nepal, while Raghavendra et al.~\cite{raghavendra2010multilingual} pointed to the lack of robust multilingual speech synthesis systems in Indian languages, emphasizing infrastructural challenges for regional screen reader development.

Researchers also explored automated solutions to tackle accessibility issues. Several approaches are proposed for alt text generation, including human-curated~\cite{zhang_ga11y_2022, gleason_making_2019}, image-search-based~\cite{guinness_caption_2018, pareddy_x-ray_2019}, and AI-based methods~\cite{wu_automatic_2017, aldahoul_exploring_2023, das2024provenance}. While human and search-based approaches emphasize contextual accuracy and reusability, AI-driven techniques offer scalable alternatives, including generating captions~\cite{shen2024altgen, moon2024sagol, hanley2021computer} or even producing the image itself with embedded descriptions~\cite{aldahoul_exploring_2023}. 
However, these methods still rely on high-quality training data and often require human oversight to ensure contextual relevance and inclusivity.
%%%%%%%%%%%%%%%%%%%%%%%%
\section{Conclusion}
%%%%%%%%%%%%%%%%%%%%%%%%
Multilingual web content is increasing in prevalence, but accessibility support lags behind, especially for non-Latin scripts. This paper introduces \tool, the first large-scale dataset of 120,000 popular websites across 12 languages that primarily use non-Latin scripts. Analysis of the \tool dataset reveals widespread issues and motivates language-aware accessibility improvements such as \testingtool,  a Google Lighthouse extension we developed, which incorporates language consistency checks into accessibility testing. We hope that our dataset, which we will open source, will spark the community's interest to further measure the implications of an increasingly multilingual web.

\bibliographystyle{ACM-Reference-Format}
\bibliography{sample-base}\

\appendix
%%%%%%%%%%%%%%%%%%%%%%%%%%%%%%%%%%
% \section{Appendix}
%%%%%%%%%%%%%%%%%%%%%%%%%%%%%%%%%%
\section{Ethics}
This work does not raise any ethical issues.

\section{Data and Tool Availability}
We have open-sourced both the \tool dataset and \testingtool. \testingtool includes detailed documentation and a \texttt{README} file explaining how to use it and how to extend it with custom accessibility tests.
We have also created an interactive website for \tool, where users can explore the dataset in greater detail, including language distribution across individual websites, with sampling and filtering options.

The dataset and testing tool are available at: 

\url{https://anonymous.4open.science/r/LangCrux-F68F/} \\
The interactive website is available at: 

\url{https://anonymous.4open.science/w/LangCrux-F68F/}

\section{Distribution of Website Rankings}
\begin{figure}[ht]
    \centering
    \includegraphics[width=\linewidth]{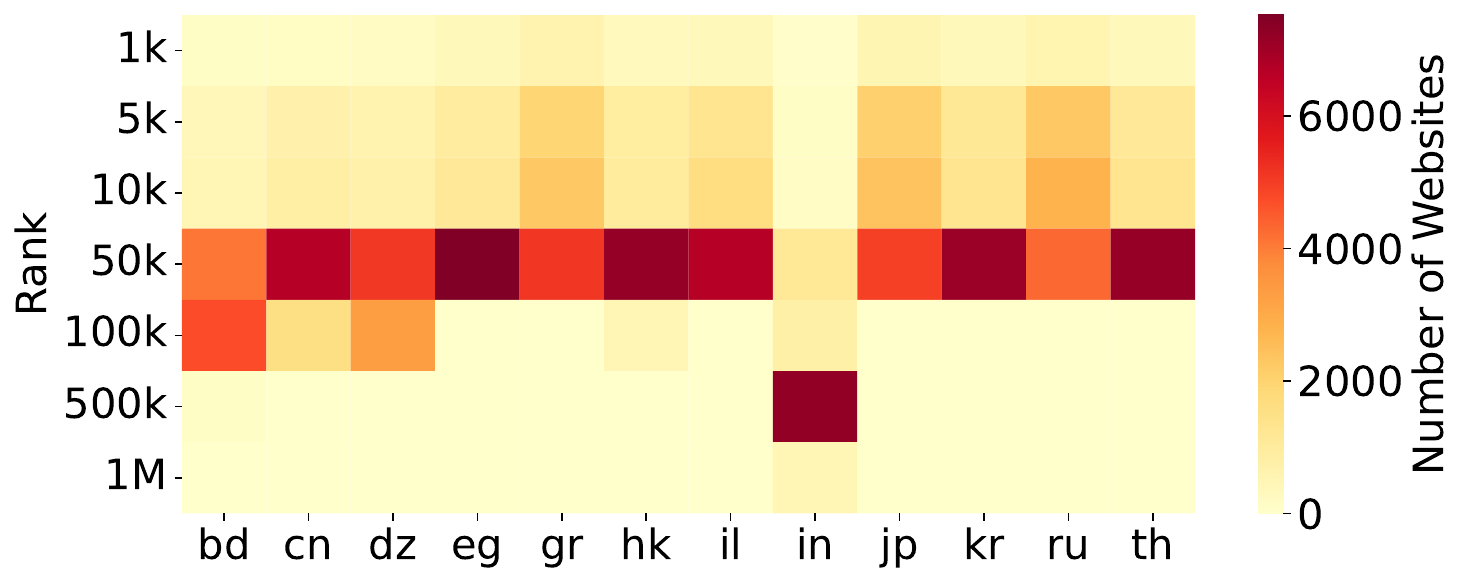}
    \caption{Distribution of website rankings across countries in \tool. Each cell indicates the number of websites within a specific global rank range for a given country. Most countries have websites concentrated in the top 50,000 ranks, while India shows a broader distribution extending toward the 1 million rank range.}
    \label{fig:heatmap_rank_distribution}
\end{figure}

Figure~\ref{fig:heatmap_rank_distribution} displays the distribution of website rankings across different countries in \tool. The rankings are based on CrUX (Chrome User Experience) data. Most countries have the majority of their websites ranked within the top 50{,}000 rank, indicating high visibility and usage. However, a notable exception is India, where the rankings tend to be significantly lower, often reaching into the 1 million range.

\section{Evaluation of Accessibility Elements Using Lighthouse}
\label{subsec:appendix-lighthouseevalution}
Table~\ref{tab:lighthouse-results} summarizes the results of our Lighthouse tests for individual accessibility elements discussed in Section~\cref{subsec:element_selection}. To understand how Lighthouse responds to different accessibility conditions, we created isolated test pages, each containing only a single target element. For each element, we evaluated three scenarios: the element being completely missing, present but with an empty value, and present with content in a different language. The table reports whether Lighthouse flagged the element as failing or passing in each scenario.

\begin{table}[h]
\centering
\caption{Lighthouse test results for individual accessibility elements under different conditions. A \cmark\ indicates the test passed, while a \xmark\ indicates the test failed.}

\label{tab:lighthouse-results}
\begin{adjustbox}{width=\linewidth}
\begin{tabular}{cccc}
\hline
\textbf{Accessibility Rule} & \textbf{Missing Element} & \textbf{Empty Value} & \textbf{Incorrect Language} \\
\hline
button-name       & \xmark & \cmark & \cmark \\
document-title    & \cmark & \xmark & \cmark \\
frame-title       & \xmark & \xmark & \cmark \\
image-alt         & \xmark & \cmark & \cmark \\
input-button-name & \cmark & \xmark & \cmark \\
input-image-alt   & \xmark & \xmark & \cmark \\
label             & \cmark & \cmark & \cmark \\
link-name         & \xmark & \xmark & \cmark \\
object-alt        & \xmark & \xmark & \cmark \\
select-name       & \xmark & \xmark & \cmark \\
summary-name      & \cmark & \cmark & \cmark \\
svg-img-alt       & \cmark & \cmark & \cmark \\
\hline
\end{tabular}
\end{adjustbox}
\end{table}

\section{Examples of Extreme Accessibility Text Values}
\label{appendix:outlier-examples}

Table~\ref{tab:long_alt_texts} shows examples of image alt texts that exceed 1000 characters. These values were extracted from real-world webpages from Bangladesh, India, Japan, Greece, and Thailand, and are shown along with the corresponding source URLs. The examples illustrate cases where accessibility attributes contain unusually long descriptive content, often including entire paragraphs or embedded metadata.
\begin{table*}[t]
\centering
\renewcommand{\arraystretch}{1.6}
\setlength{\tabcolsep}{10pt}
\begin{tabularx}{\textwidth}{|m{0.55\textwidth}|X|}
\hline
\textbf{Image Alt Text} & \textbf{Source URL} \\
\hline
\includegraphics[width=0.48\textwidth]{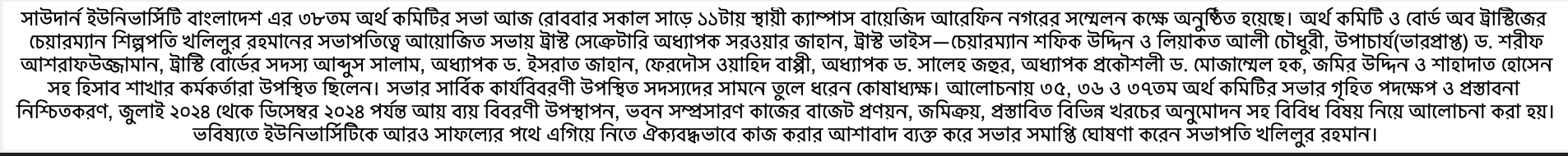} & \url{https://www.shiksharalo.net} \\
\hline
\includegraphics[width=0.5\textwidth]{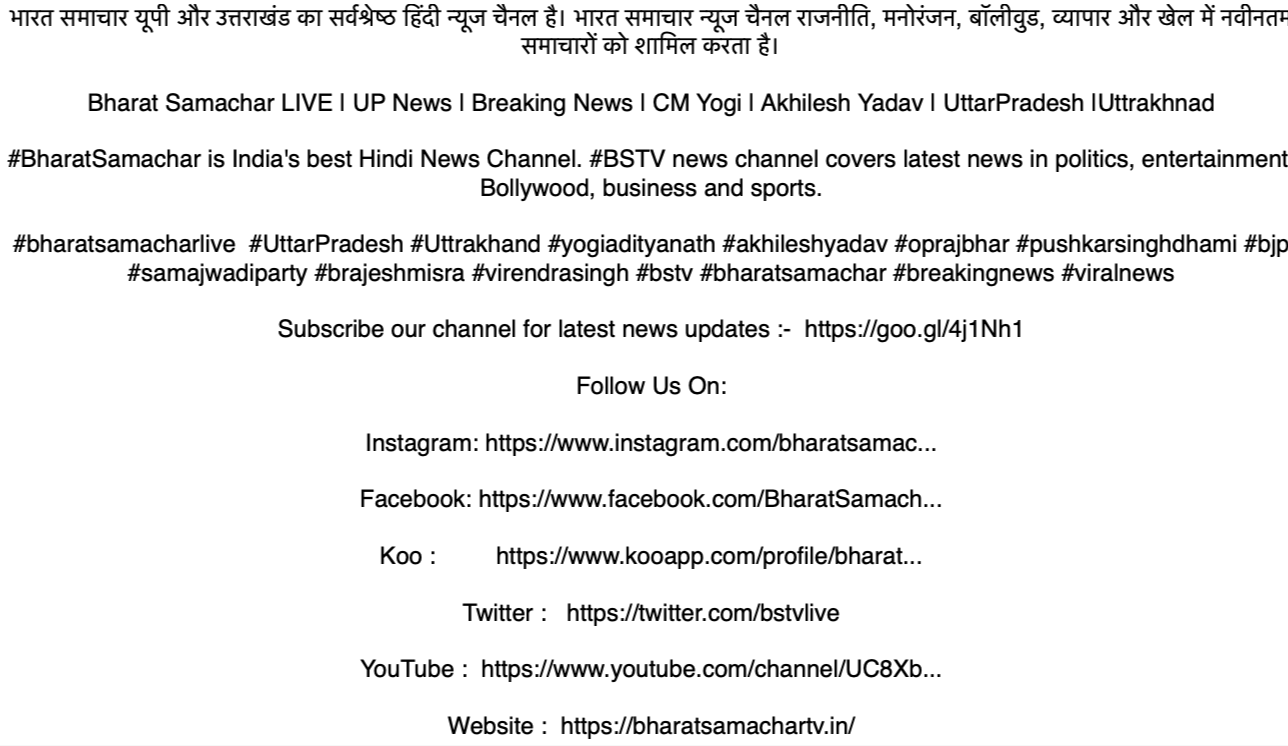} & \url{https://bharatsamachartv.in} \\
\hline
\includegraphics[width=0.52\textwidth]{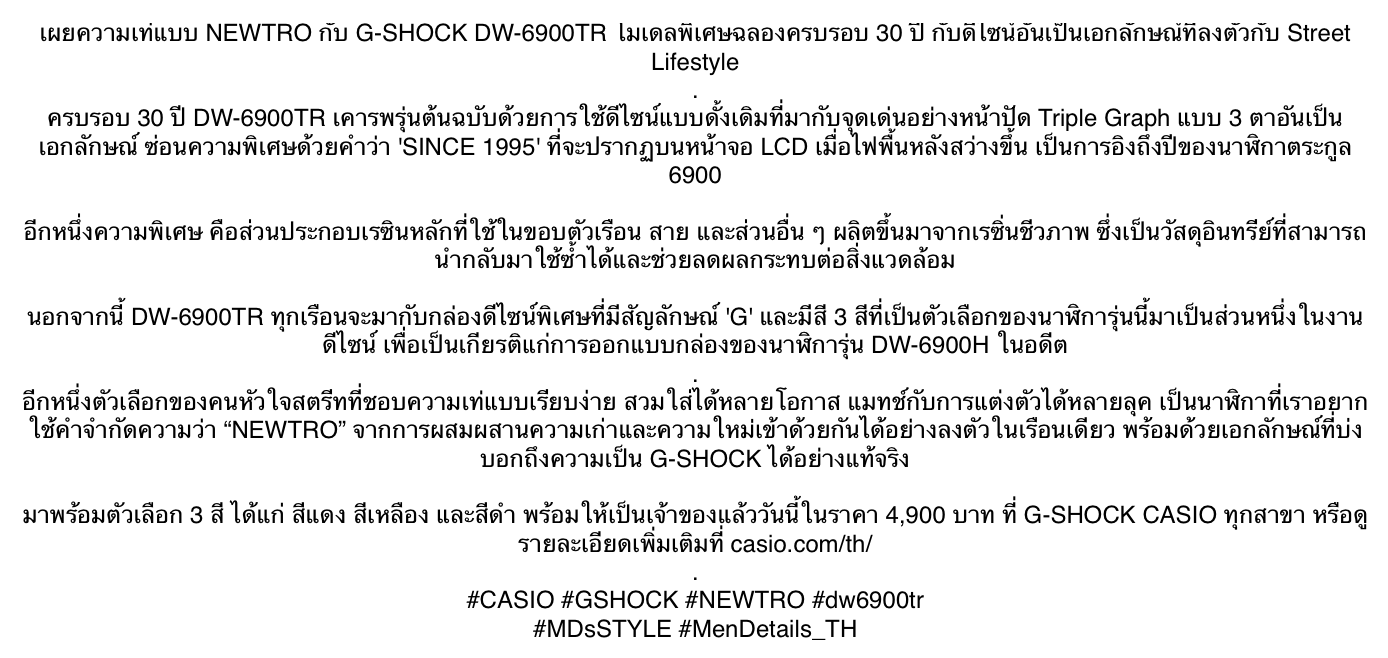} & \url{https://www.mendetails.com} \\
\hline
\includegraphics[width=0.52\textwidth]{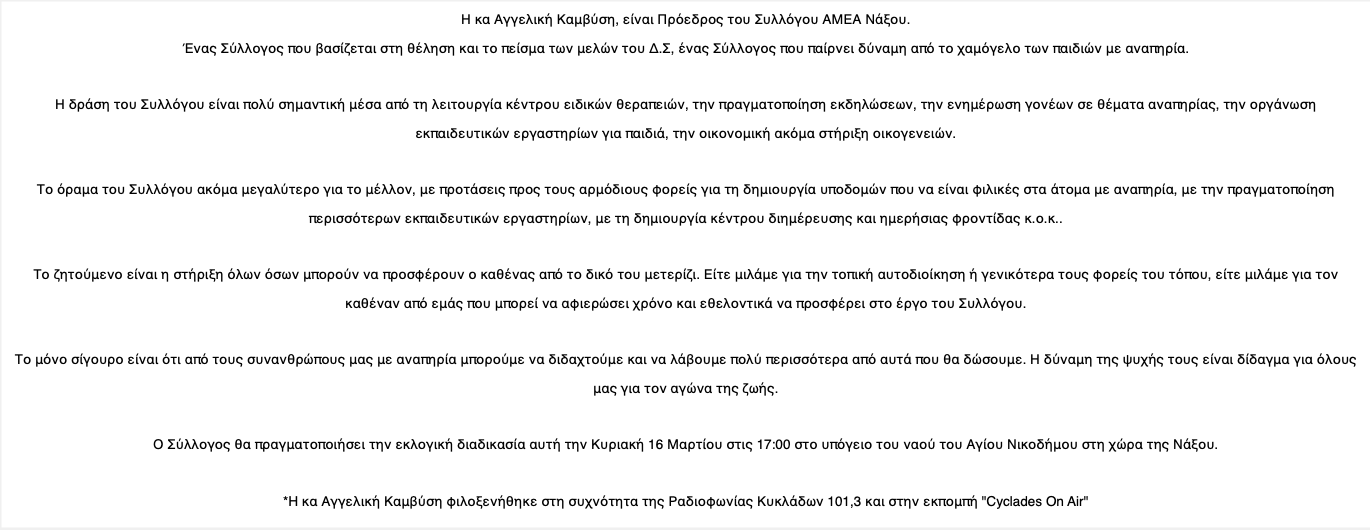} & \url{https://naxospress.gr} \\
\hline
\includegraphics[width=0.48\textwidth]{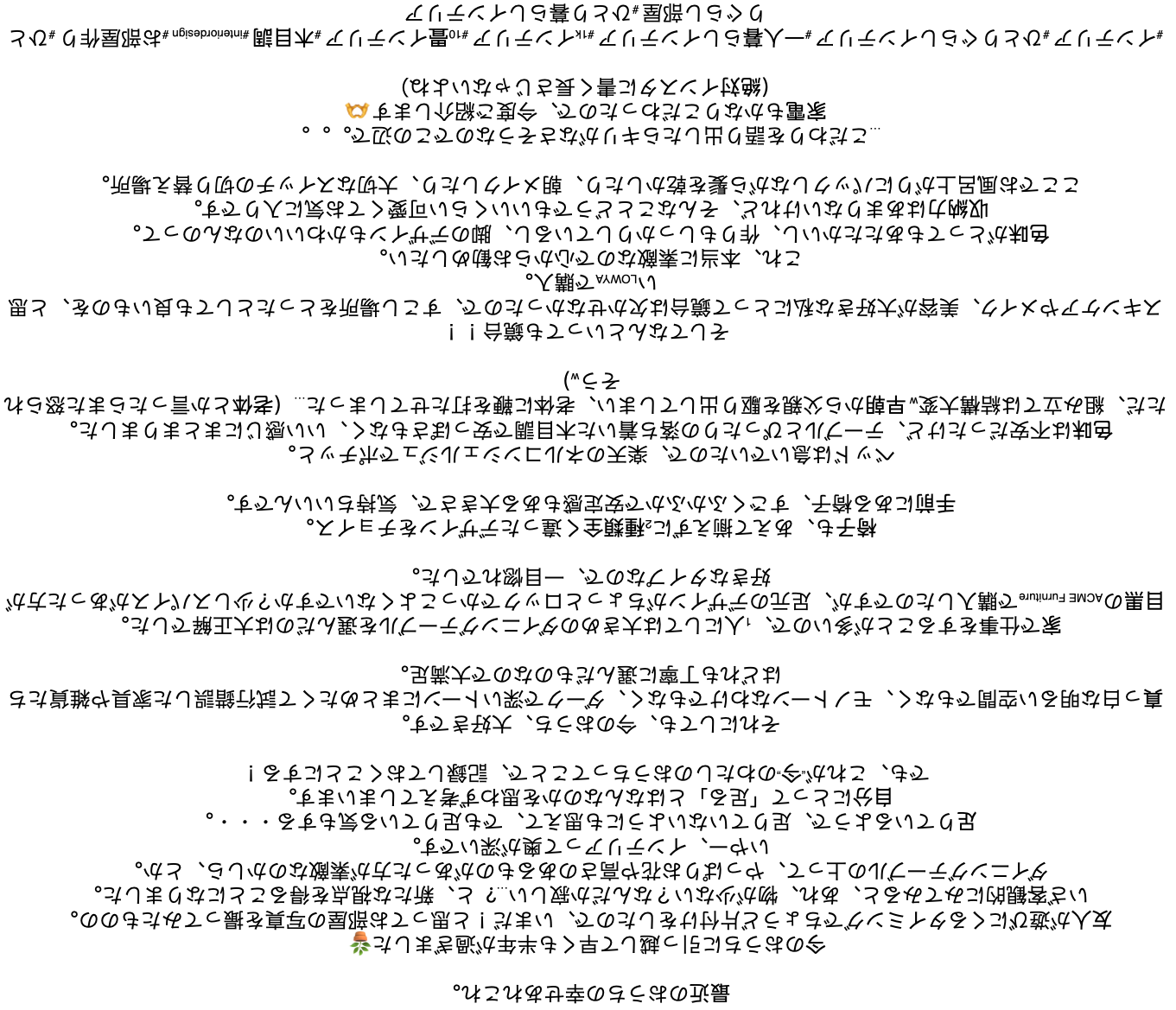} & \url{https://www.low-ya.com} \\
\hline
\end{tabularx}
\caption{Examples of image alt texts exceeding 1000 characters in length, collected from websites in Bangladesh, India, Thailand, Greece, and Japan. Each row shows the corresponding alt text and the source URL where the alt text was extracted.}
\label{tab:long_alt_texts}
\end{table*}

\section{Country-Level Scatter Plots}
\label{appendix:scatterplots}
\begin{figure*}
\centering

% Row 1
\begin{subfigure}[t]{0.31\textwidth}
\includegraphics[width=\linewidth]{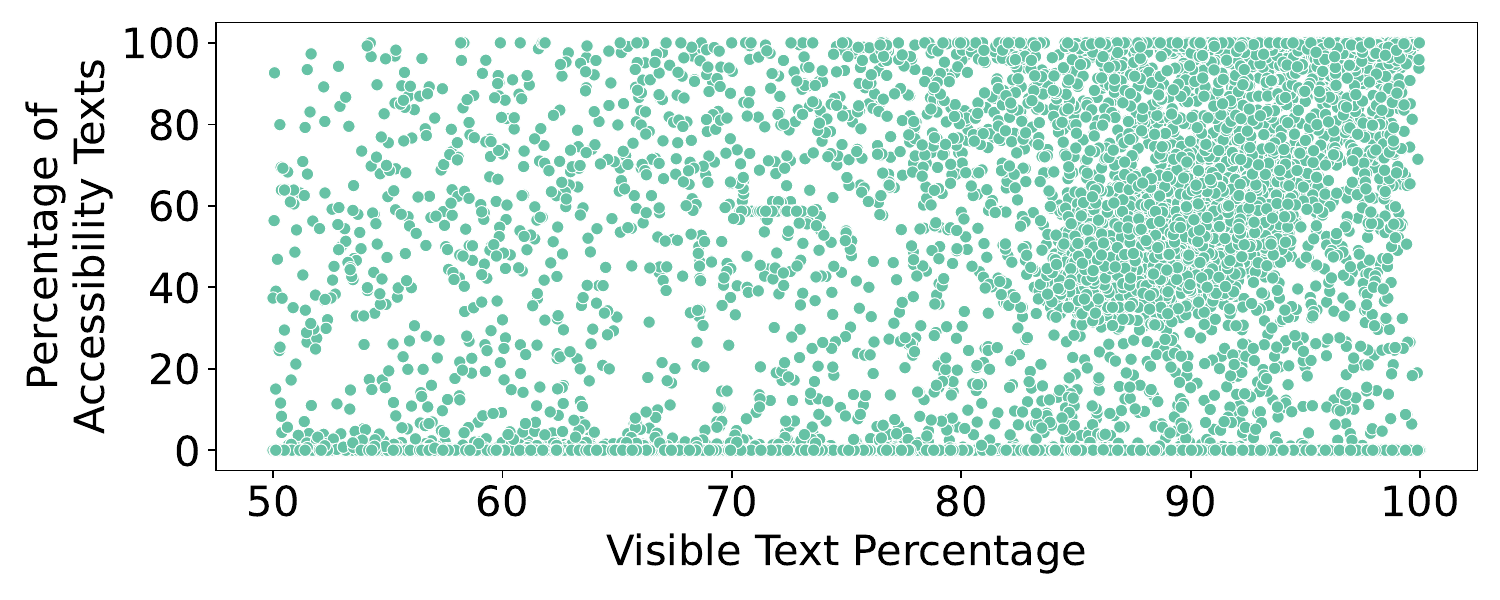}
\caption{Bangladesh}
\label{fig:scatter_bd}
\end{subfigure}
\hfill
\begin{subfigure}[t]{0.31\textwidth}
\includegraphics[width=\linewidth]{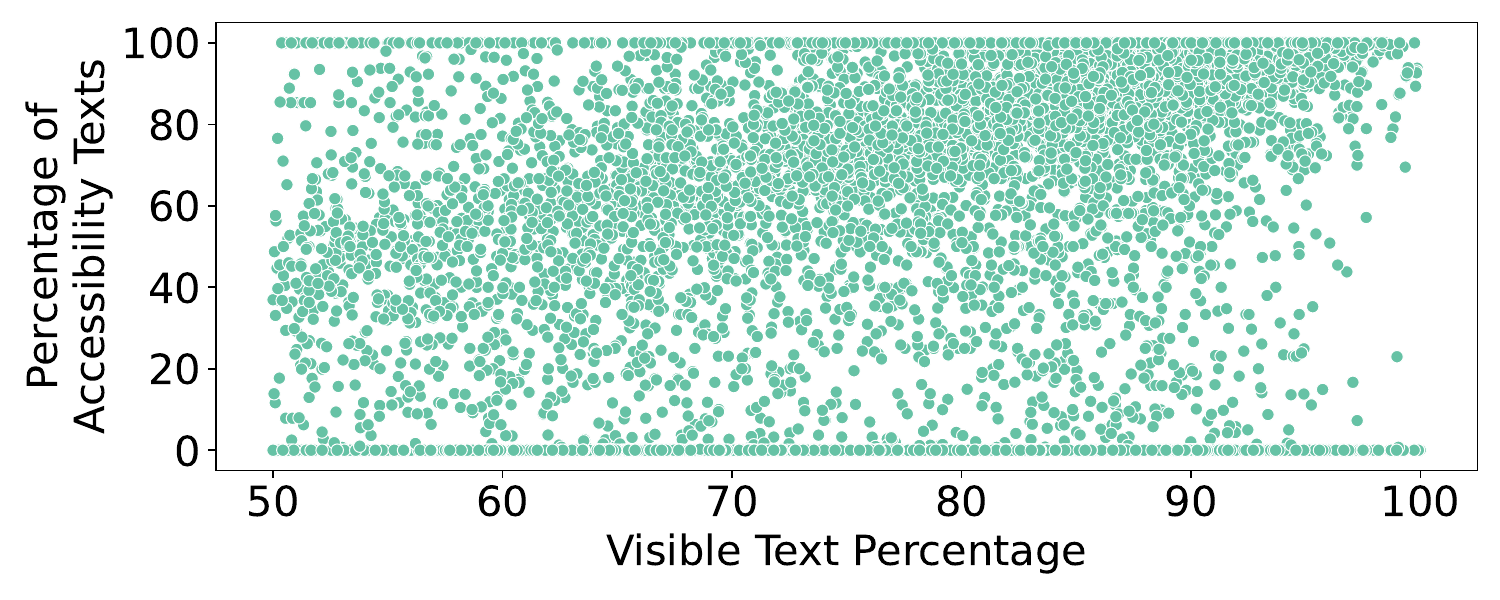}
\caption{China}
\label{fig:scatter_cn}
\end{subfigure}
\hfill
\begin{subfigure}[t]{0.31\textwidth}
\includegraphics[width=\linewidth]{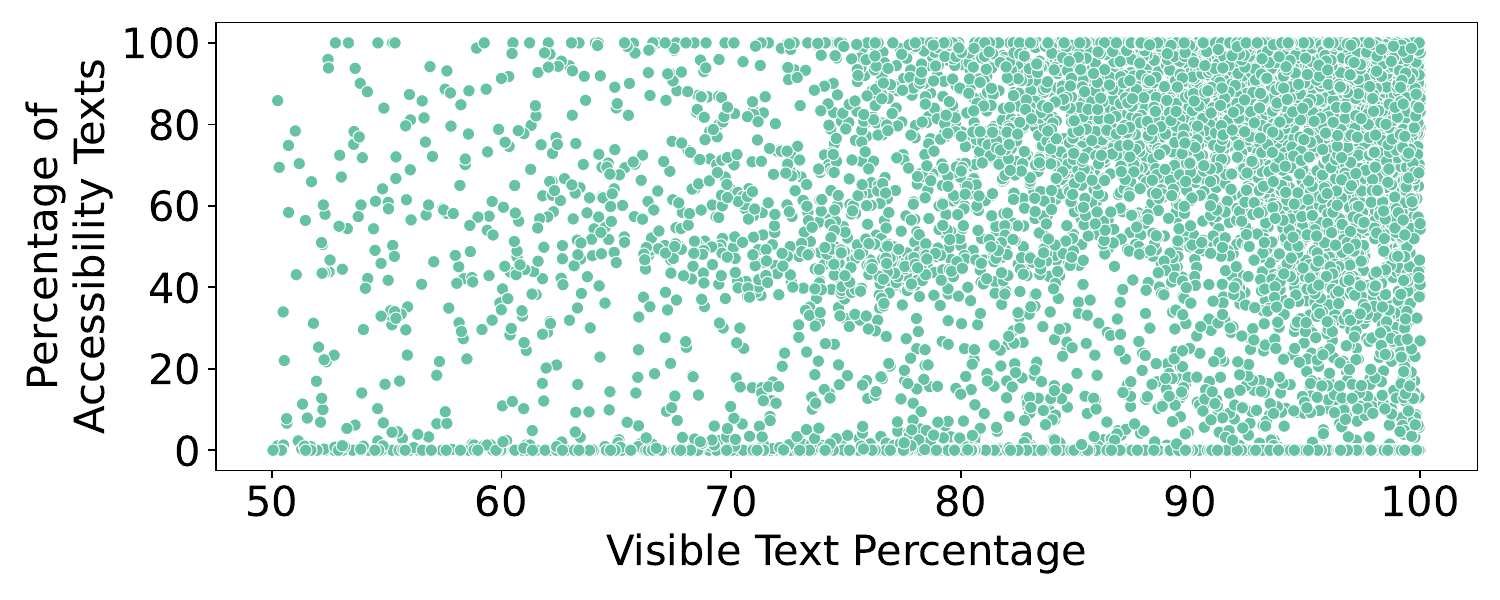}
\caption{Egypt}
\label{fig:scatter_eg}
\end{subfigure}

\vspace{1em}

% Row 2
\begin{subfigure}[t]{0.31\textwidth}
\includegraphics[width=\linewidth]{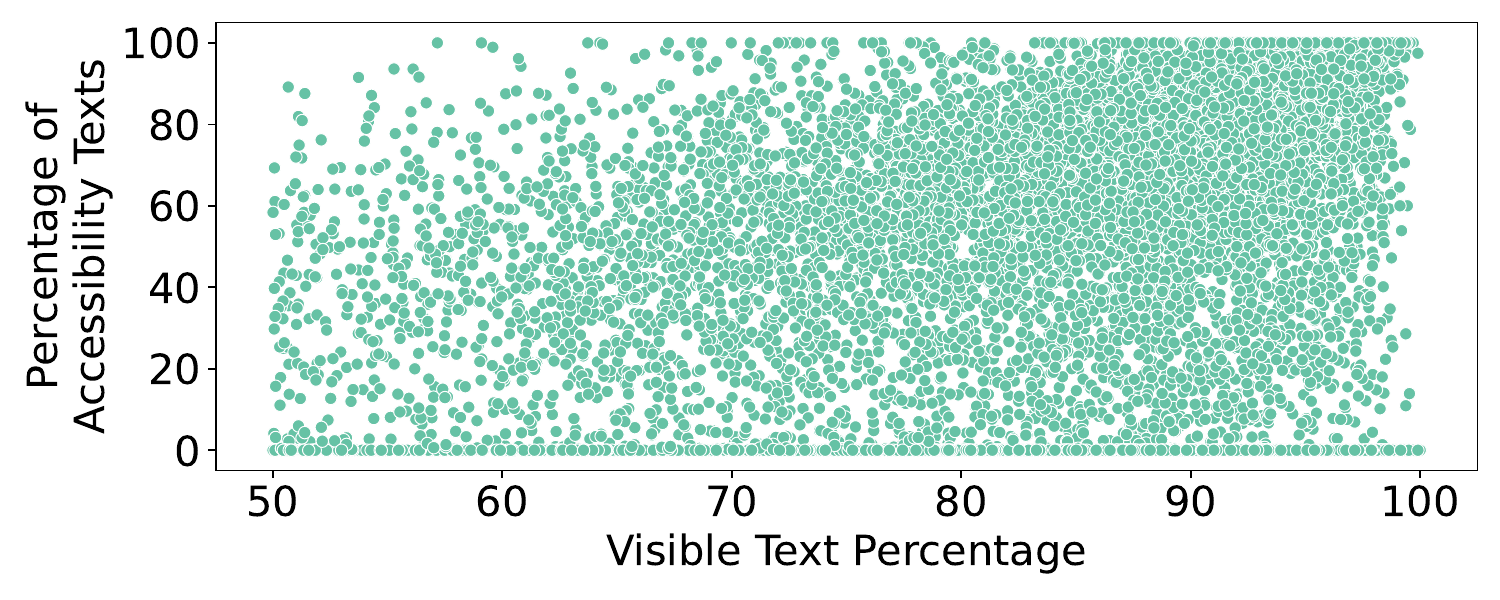}
\caption{Greece}
\label{fig:scatter_gr}
\end{subfigure}
\hfill
\begin{subfigure}[t]{0.31\textwidth}
\includegraphics[width=\linewidth]{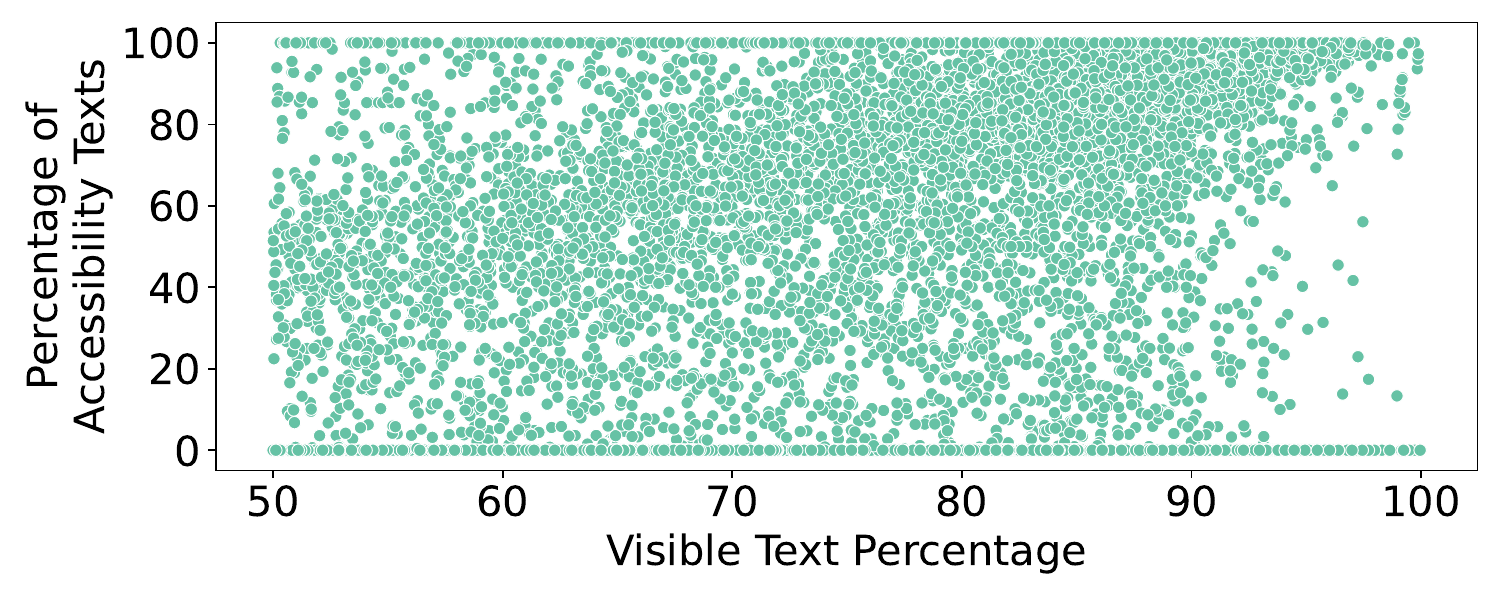}
\caption{Hong Kong}
\label{fig:scatter_hk}
\end{subfigure}
\hfill
\begin{subfigure}[t]{0.31\textwidth}
\includegraphics[width=\linewidth]{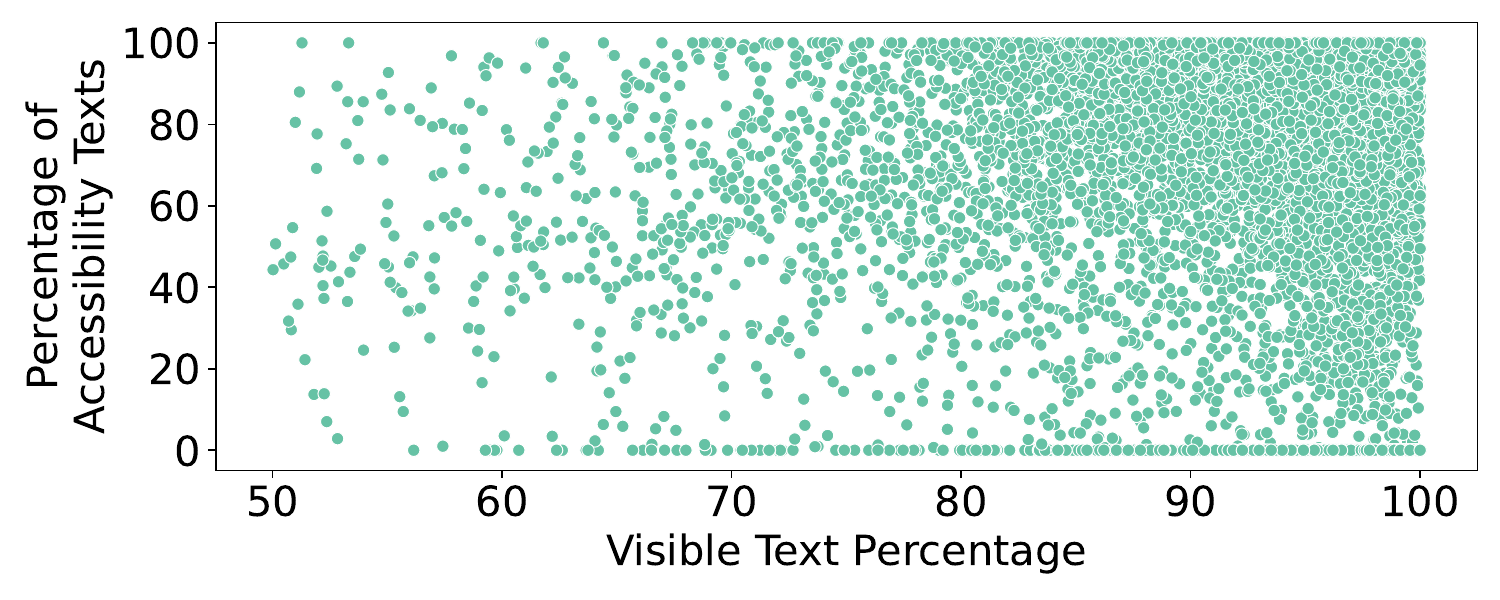}
\caption{Israel}
\label{fig:scatter_il}
\end{subfigure}

\vspace{1em}

% Row 3
\begin{subfigure}[t]{0.31\textwidth}
\includegraphics[width=\linewidth]{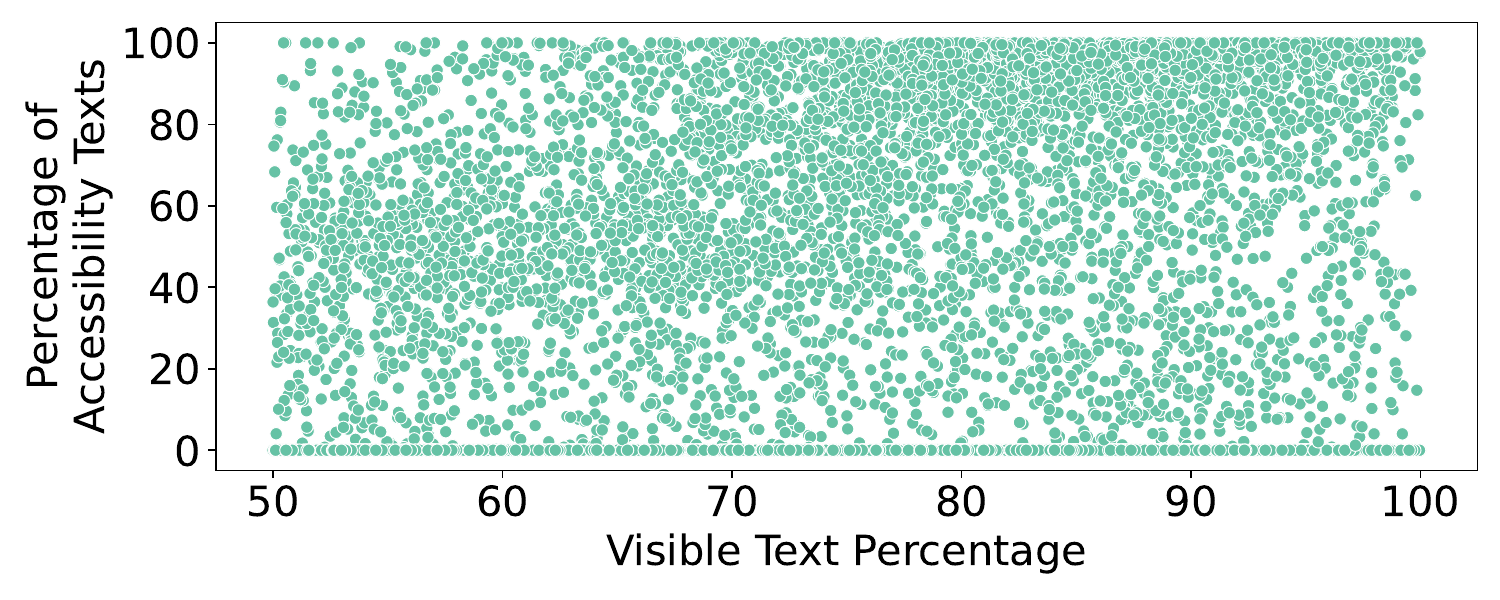}
\caption{India}
\label{fig:scatter_in}
\end{subfigure}
\hfill
\begin{subfigure}[t]{0.31\textwidth}
\includegraphics[width=\linewidth]{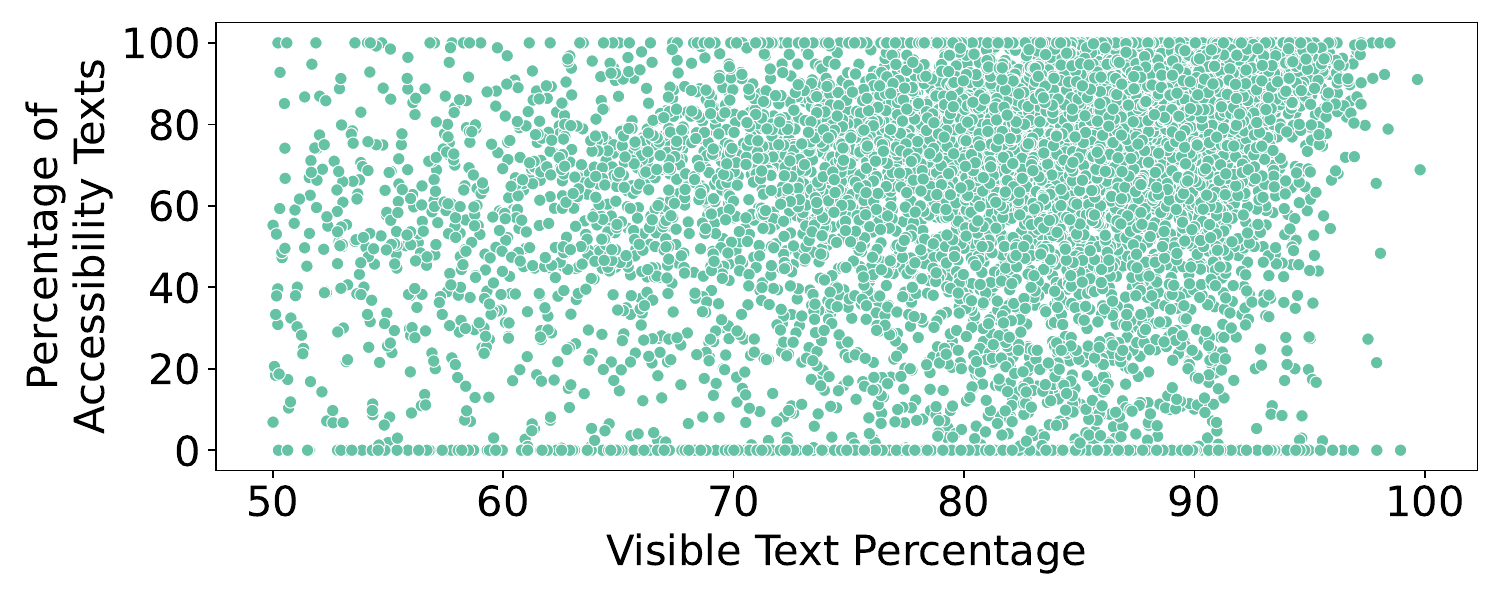}
\caption{Japan}
\label{fig:scatter_jp}
\end{subfigure}
\hfill
\begin{subfigure}[t]{0.31\textwidth}
\includegraphics[width=\linewidth]{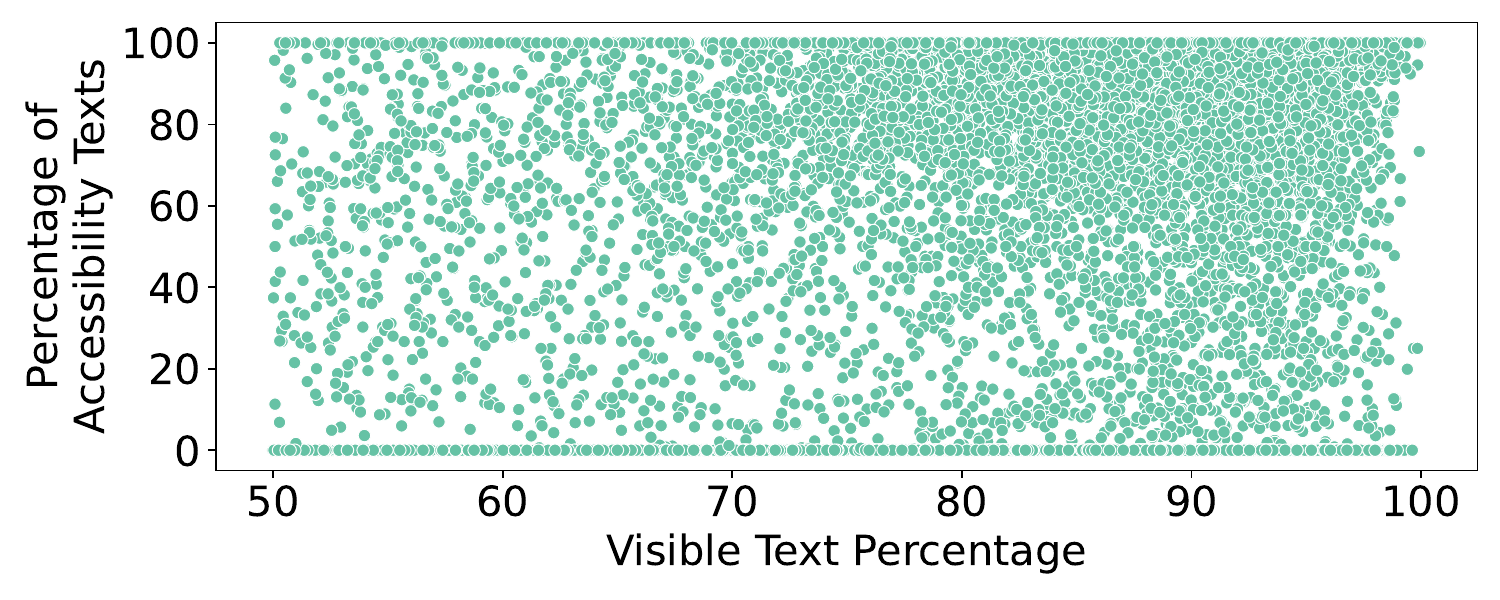}
\caption{South Korea}
\label{fig:scatter_kr}
\end{subfigure}

\vspace{1em}

% Row 4
\begin{subfigure}[t]{0.31\textwidth}
\includegraphics[width=\linewidth]{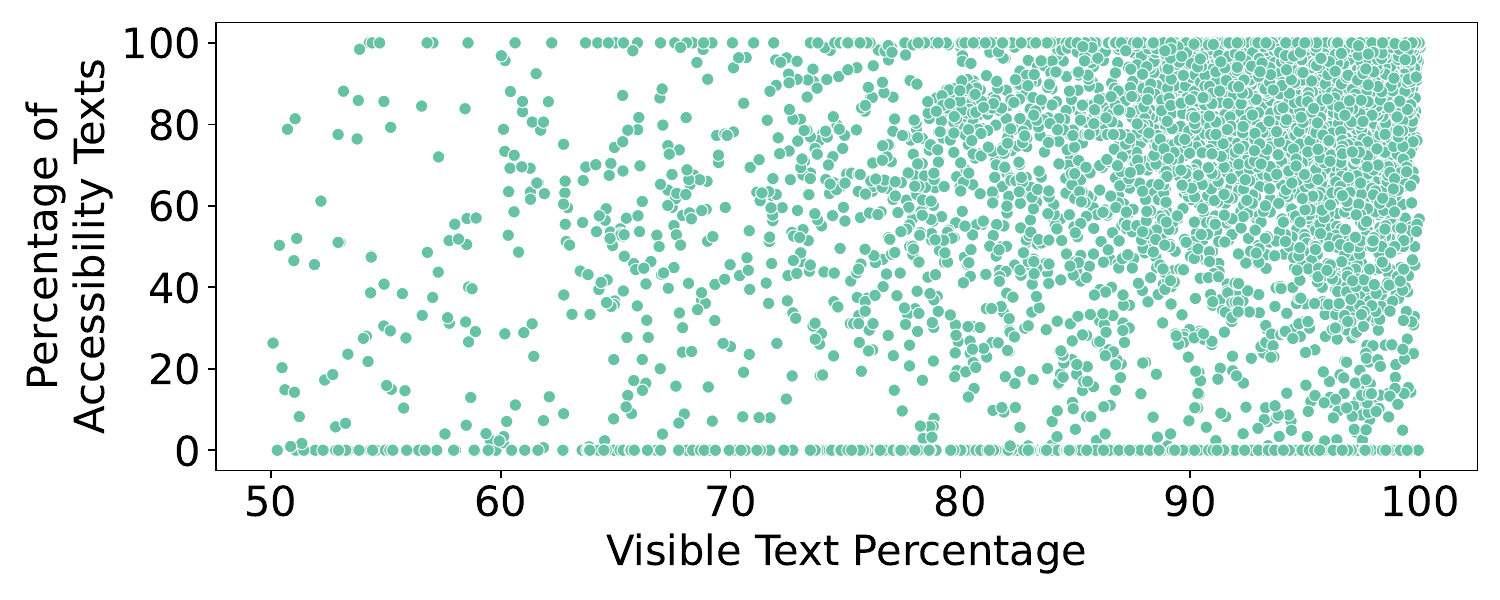}
\caption{Russia}
\label{fig:scatter_ru}
\end{subfigure}
\hfill
\begin{subfigure}[t]{0.31\textwidth}
\includegraphics[width=\linewidth]{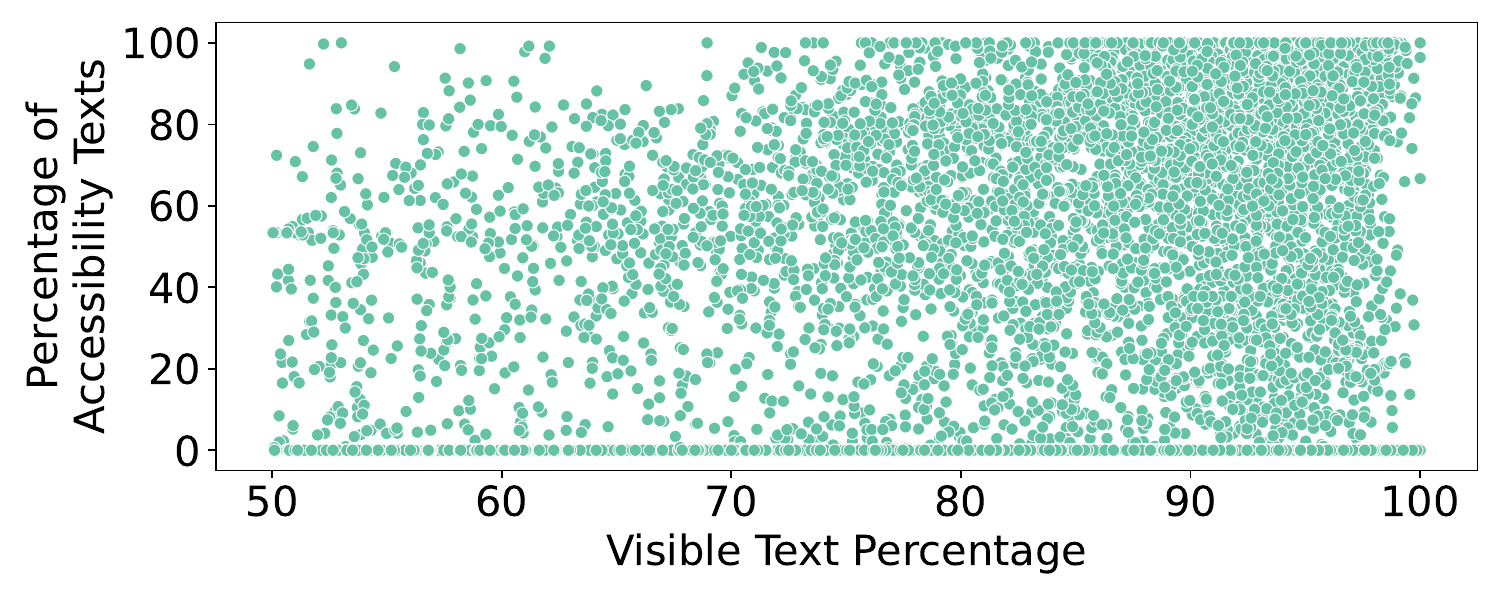}
\caption{Thailand}
\label{fig:scatter_th}
\end{subfigure}
\hfill
\begin{subfigure}[t]{0.31\textwidth}
\includegraphics[width=\linewidth]{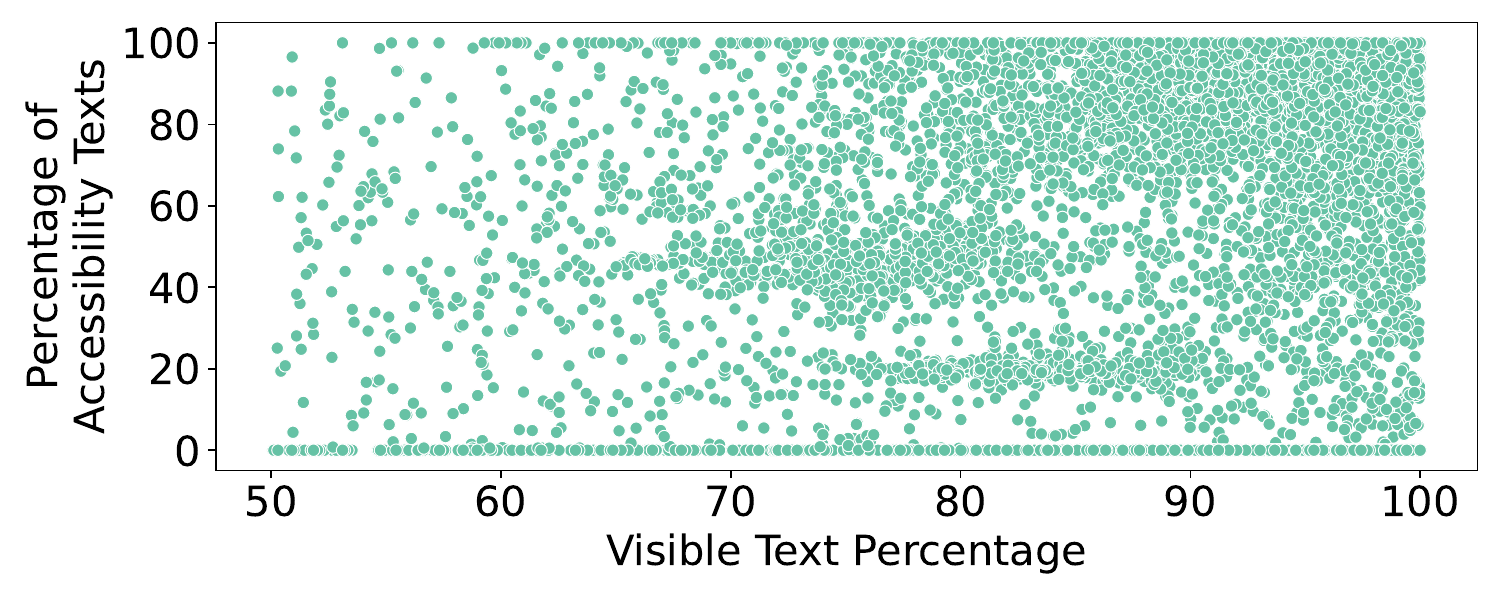}
\caption{Algeria}
\label{fig:scatter_dz}
\end{subfigure}

\caption{Scatter plot showing the percentage of native language usage in visible text versus accessibility text for websites from 12 different countries. Each point represents a single website. The plots highlight the degree of alignment or mismatch between the languages used in visible content and the corresponding accessibility attributes.}
\label{fig:lang_match_scatter}
\end{figure*}

To complement the main analysis, we include country-specific scatter plots, shown in Figure~\ref{fig:lang_match_scatter}, displaying the distribution of websites based on the percentage of native-language content in visible versus accessibility text. Each point represents one site, with the x-axis representing the share of visible content in the native language, and the y-axis representing the share of accessibility metadata in the same language. These plots offer a more detailed view of the language mismatch patterns discussed in Section~\cref{sec:res}.

As an illustration, consider the scatter plot for Thai websites, shown in Figure~\ref{fig:scatter_th}. Points near the bottom of the plot represent websites where there is almost no accessibility metadata in Thai, regardless of the language used in the visible content. A dense cluster in the bottom right corner indicates websites where the visible content is almost entirely in Thai, but the accessibility text includes little or no Thai. In contrast, points in the top right corner represent websites where both the visible and accessibility content are predominantly in Thai, indicating consistent language use across both types of content.

\section{Element-Level Filtering Analysis}
\label{appendix:element_filtering}

\begin{figure}[H]
  \centering
  \includegraphics[width=\linewidth]{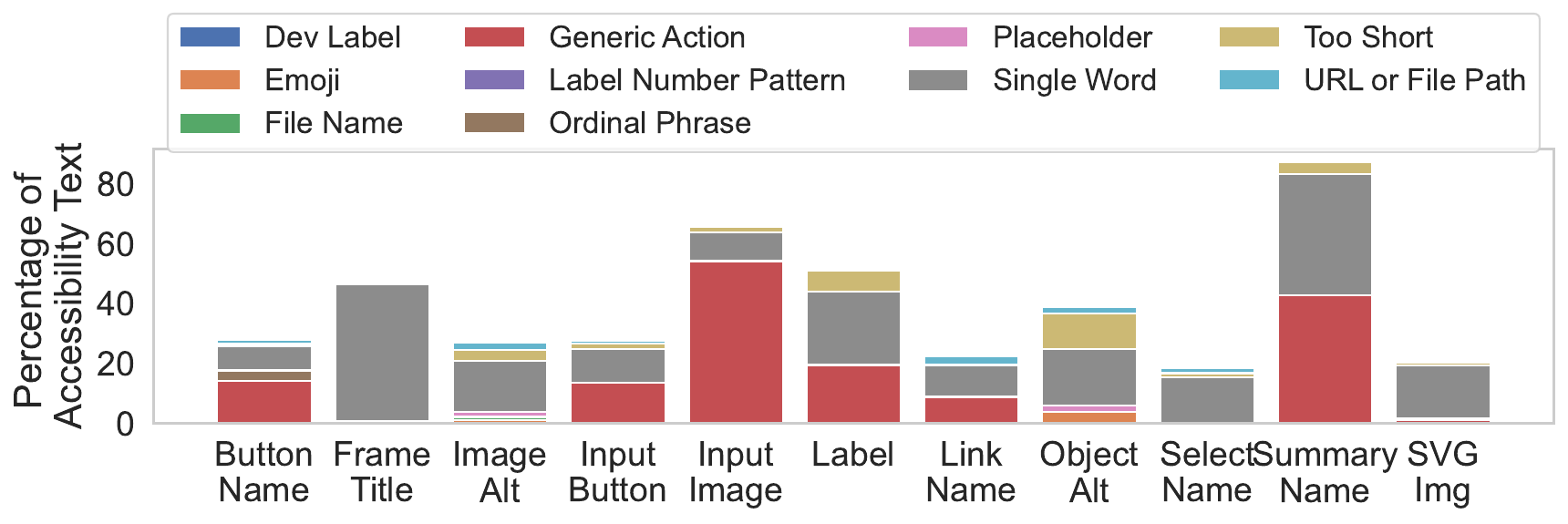}
  \caption{Breakdown of filtered uninformative accessibility text by HTML element and category.}
  \label{fig:filtered_reason_element}
\end{figure}

Figure~\ref{fig:filtered_reason_element} shows the distribution of uninformative accessibility text by HTML element. Generic action labels are especially common in \texttt{<button>} (14.2\%) and \texttt{<input>} buttons (13.5\%), indicating vague, non-descriptive usage. Single-word labels are the most prevalent issue overall, notably in \texttt{<label>} (24.4\%), \texttt{<image-alt>} (17.1\%), and \texttt{<select>} (15.3\%) elements. These short, generic strings often fail to provide meaningful context. \texttt{<summary>} elements show both high generic action (42.9\%) and single-word rates (40.5\%), highlighting minimal semantic value. These trends suggest a need for deeper evaluation of accessibility text beyond its presence.

\section{Filtering Uninformative Accessibility Text}
\label{appendix:filtering}

To assess the informativeness of accessibility text (e.g., \texttt{alt}, \texttt{aria-label}, or \texttt{label} attributes), we apply a rule-based filtering pipeline to discard generic or low-quality entries. Below, we outline each discard category, its rationale, and representative examples.

\begin{itemize}
    \item \textbf{Emoji:} Emoji are discarded because screen readers often fail to interpret them reliably or skip them altogether, making them unsuitable for conveying meaningful accessible content.\\
    \emph{Example:} [emoji character removed for compatibility]

    \item \textbf{Too Short:} Texts below a language-specific character threshold are considered too short to be useful. For CJK (Chinese, Japanese, Korean) scripts, the limit is 1 character; for others, it is 3 characters.\\
    \emph{Example:} "go", \begin{CJK*}{UTF8}{gbsn}"图"\end{CJK*}

    \item \textbf{File Name:} Strings that appear to be image or asset file names (e.g., ending in .jpg, .png, .svg) are removed.\\
    \emph{Example:} "banner\_img123.jpg"

    \item \textbf{URL or File Path:} URLs or file system paths are excluded as they are not meaningful for screen reader users.\\
    \emph{Example:} "https://example.com/image.png", \\"/assets/img/logo.svg"

    \item \textbf{Generic Action:} Common UI actions (e.g., “close”, “search”) in multiple languages are filtered if used alone without context.\\
    \emph{Example:} "search", \begin{CJK}{UTF8}{mj}"닫기"\end{CJK} (Korean for "close")

    \item \textbf{Placeholder:} Generic placeholders for images or UI components, such as "image", "icon", or "button", are removed. These include translations in various languages.\\
    \emph{Example:} "icon", \begin{CJK*}{UTF8}{gbsn}"图像"\end{CJK*} (Chinese for "image")

    \item \textbf{Dev Label:} Developer-generated IDs or component labels (e.g., "navbar-toggle", "carousel1") are excluded.\\
    \emph{Example:} "btn-submit", "nav\_menu"

    \item \textbf{Label Number Pattern:} Common patterns like "image 1", "button 2", etc., are discarded as they typically lack descriptive value.\\
    \emph{Example:} "slide 3", "figure 5"

    \item \textbf{Single Word:} For non-CJK scripts, single-word entries are filtered unless they appear to carry descriptive meaning.\\
    \emph{Example:} "photo", "submit"

    \item \textbf{Mixed Alnum:} Strings with alphanumeric IDs are typically programmatic or internal references and are removed.\\
    \emph{Example:} "img123", "icon2"

    \item \textbf{Ordinal Phrase:} Numeric phrases like "3 of 5" are common in pagination or sliders and provide limited context.\\
    \emph{Example:} "2 of 10", "1 of 3"
\end{itemize}

Text that does not match any of the above patterns is retained and considered \textbf{useful} for accessibility analysis. This filtering process helps distinguish meaningful metadata from boilerplate or autogenerated labels, enabling more accurate assessments of multilingual accessibility practices.

\section{Examples of Mismatch Between Visible and Accessibility Text}
\label{appendix:mismatch_examples}

Table~\ref{tab:mismatch_examples} illustrates examples of accessibility mismatches on six websites across Bangladesh, India, Thailand, Egypt, China, and Hong Kong. Each cell consists of an image from the website, the corresponding URL, and the alt text or accessibility description associated with that image.
The examples highlight cases where website content is presented in the native language, yet the accessibility descriptions, such as alt texts, are written in English. This language inconsistency creates confusion for screen reader users and demonstrates the need for language-aligned accessibility practices. 
\begin{table*}[t]
\centering
\renewcommand{\arraystretch}{2}
\setlength{\tabcolsep}{6pt}
\begin{tabular}{| >{\centering\arraybackslash}m{8.5cm} | >{\centering\arraybackslash}m{8.5cm} |}
\hline

\begin{tabular}{m{7cm}}
\centering
\includegraphics[width=4cm]{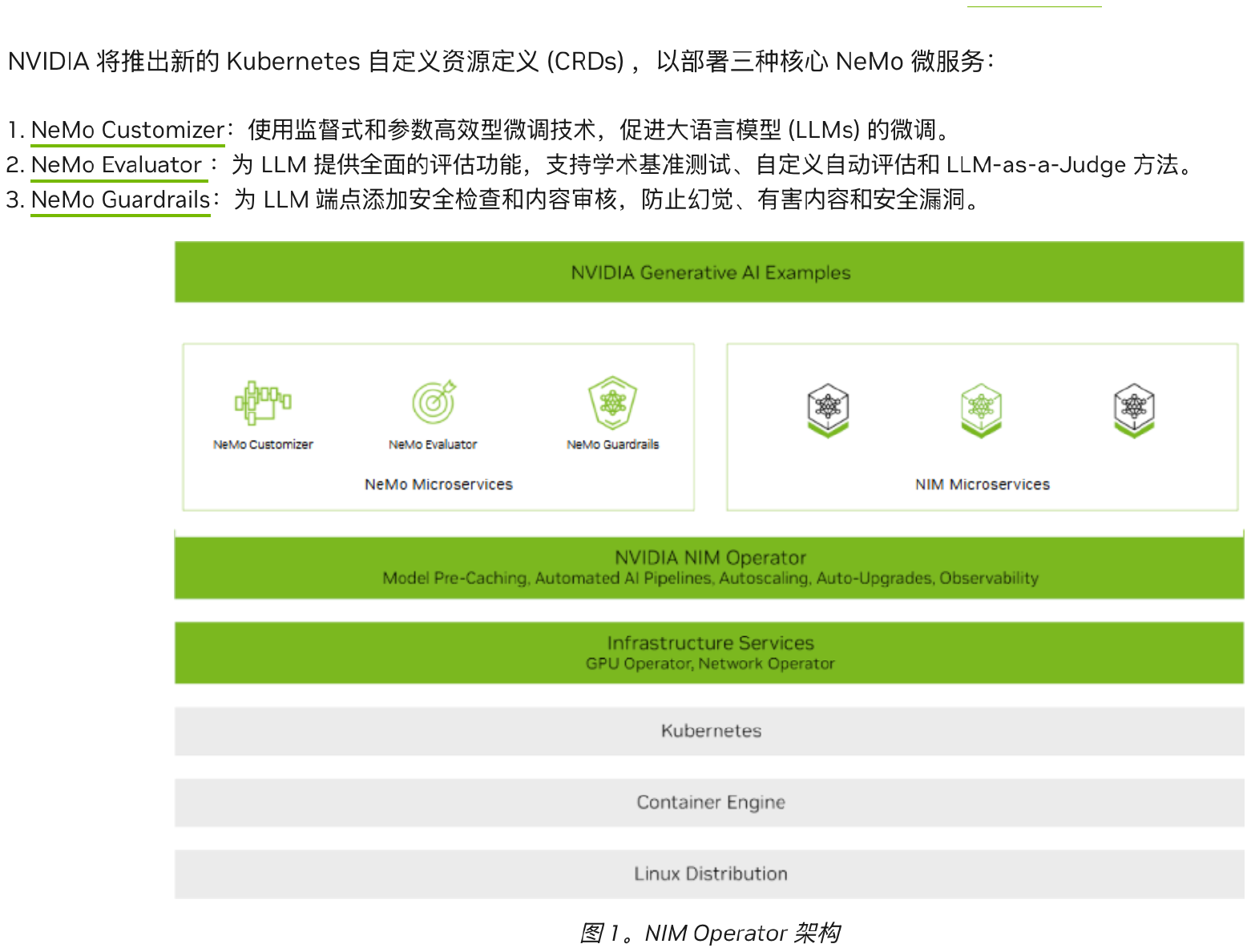} \\[0.5em]
\href{https://developer.nvidia.cn/zh-cn/blog/nvidia-nim-operator-2-0-boosts-ai-deployment-with-nvidia-nemo-microservices-support/}
{\parbox{7cm}{\small \textbf{Link:} https://developer.nvidia.cn/zh-cn/blog/nvidia-nim-operator-2-0-boosts-ai-deployment-with-nvidia-nemo-microservices-support/}} \\[0.5em]
{\parbox{7cm}{\small \textbf{Alt text:} ``The image depicts a stack diagram highlighting NVIDIA NIM Operator, a Kubernetes Operator that is designed to facilitate the deployment, management, and scaling of NVIDIA NIM microservices on Kubernetes clusters.''}}\\[0.5em]
\end{tabular}
&
\begin{tabular}{m{7cm}}
\centering
\includegraphics[width=4cm]{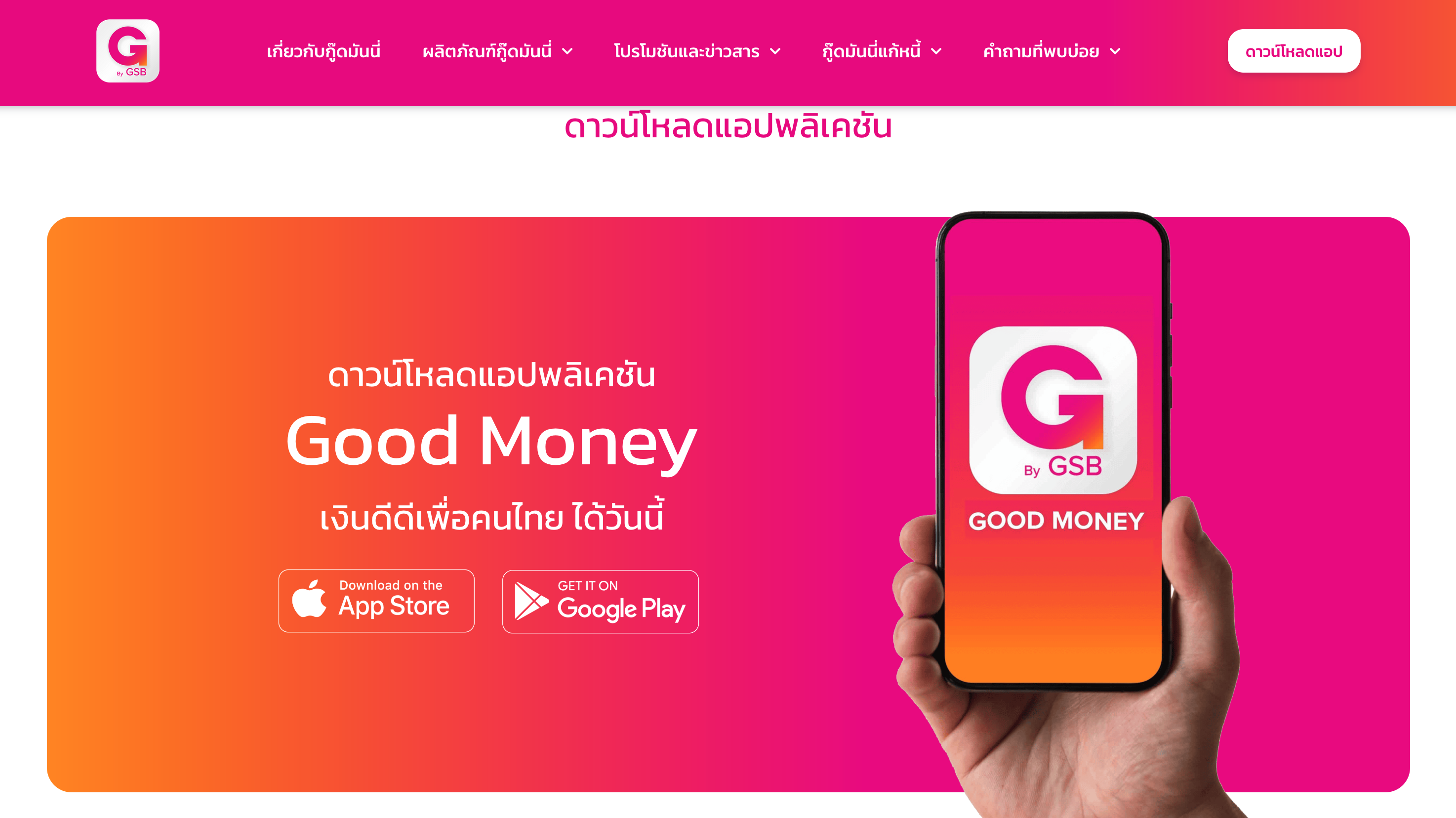} \\[0.5em]

\href{https://goodmoneybygsb.com}
{\parbox{7cm}{\small \textbf{Link:} https://goodmoneybygsb.com}} \\[0.5em]
{\parbox{7cm}{\small \textbf{Alt text:}\small ``A hand is holding a smartphone displaying the "GOOD MONEY" app by GSB. The screen shows the app's logo, featuring a white "G" on a gradient background transitioning from pink to orange.''}}\\[0.5em]
\end{tabular}
\\
\hline

\begin{tabular}{m{7cm}}
\centering
\includegraphics[width=4cm]{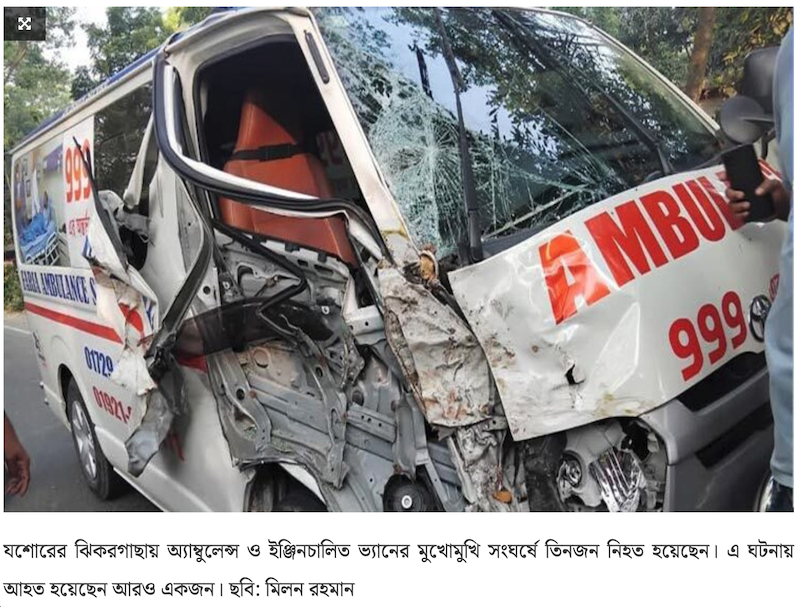} \\[0.5em]

\href{https://www.jagonews24.com/photo/bangladesh/news/12915/}
{\parbox{7cm}{\small \textbf{Link:} https://www.jagonews24.com/photo/bangladesh/ news/12915/}} \\[0.5em]
{\parbox{7cm}{\small \textbf{Alt text:}\small ``Three people were killed in a head-on collision between an ambulance and an engine-powered van in Jhikargachha, Jessore. Another person was injured in the incident. Photo: Milan Rahman''}}\\[0.5em]
\end{tabular}
&
\begin{tabular}{m{7cm}}
\centering
\includegraphics[width=4cm]{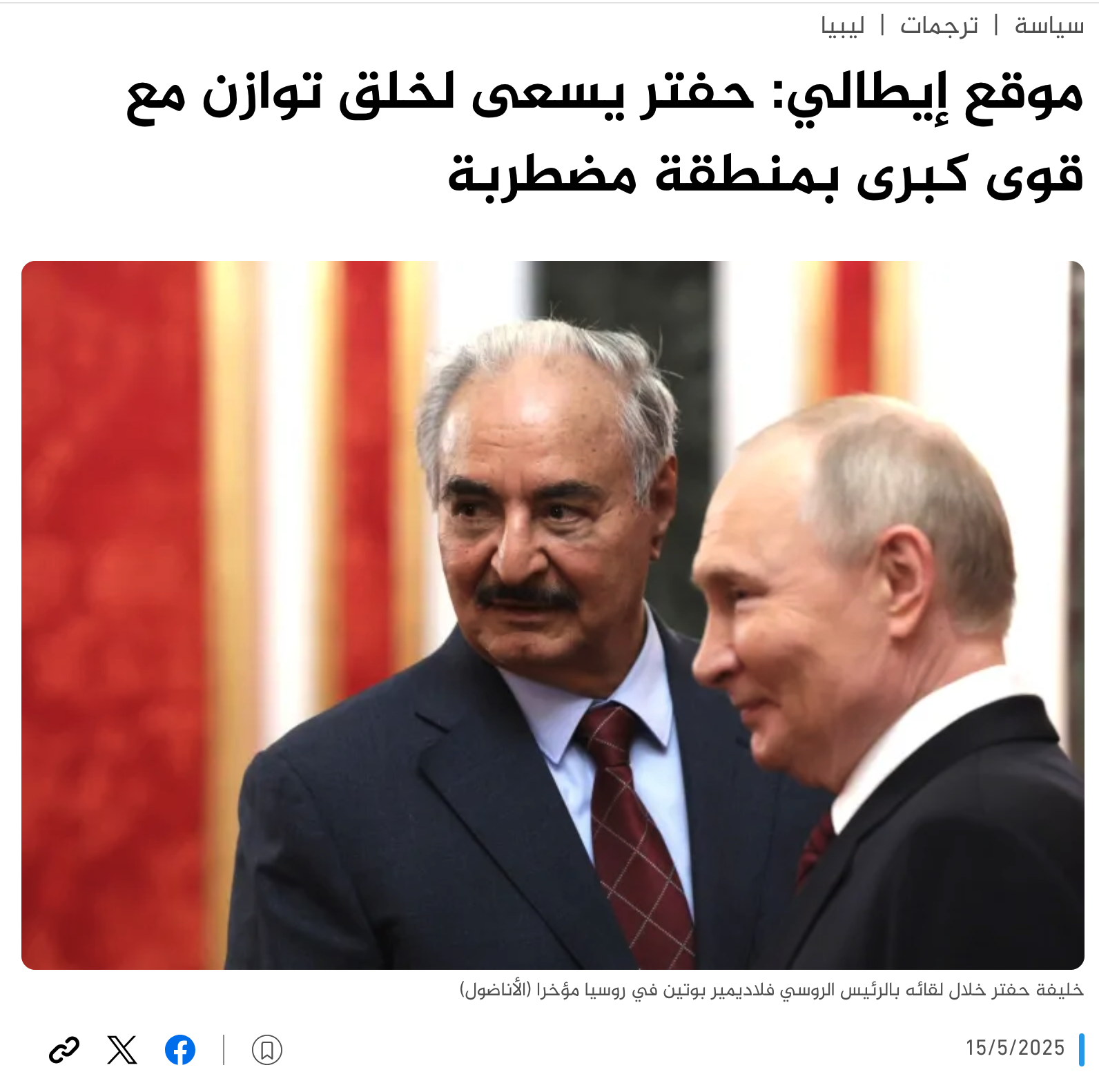} \\[0.5em]
 \href{https://www.ajnet.me/news/2025/5/15/%D8%B9%D8%A7%D8%AC%D9%84-%D8%AD%D9%85%D8%A7%D8%B3-%D9%86%D8%AF%D8%B9%D9%88-%D9%84%D8%AA%D9%83%D9%88%D9%86-%D8%A3%D9%8A%D8%A7%D9%85-%D8%A7%D9%84%D8%AC%D9%85%D8%B9%D8%A9}
{\parbox{7cm}{\small \textbf{Link:} https://www.ajnet.me}} \\[0.5em]
{\parbox{7cm}{\small \textbf{Alt text:} \small ``Vladimir Putin - Khalifa Haftar meeting in Moscow Vladimir Putin - Khalifa Haftar meeting in Moscow- - MOSCOW, RUSSIA - MAY 10: (----EDITORIAL USE ONLY - MANDATORY CREDIT - 'KREMLIN PRESS SERVICE / HANDOUT' - NO MARKETING NO ADVERTISING CAMPAIGNS - DISTRIBUTED AS A SERVICE TO CLIENTS----) President of Russia Vladimir Putin (R) meets with Khalifa Haftar, the leader of the armed forces in the east of the country in Moscow, Russia on May 10, 2025. DATE 11/05/2025 SIZE x Country Rusya SOURCE Anadolu/Kremlin Press Service.''}}\\[0.5em]
\end{tabular}
\\
\hline

\begin{tabular}{m{7cm}}
\centering
\includegraphics[width=4cm]{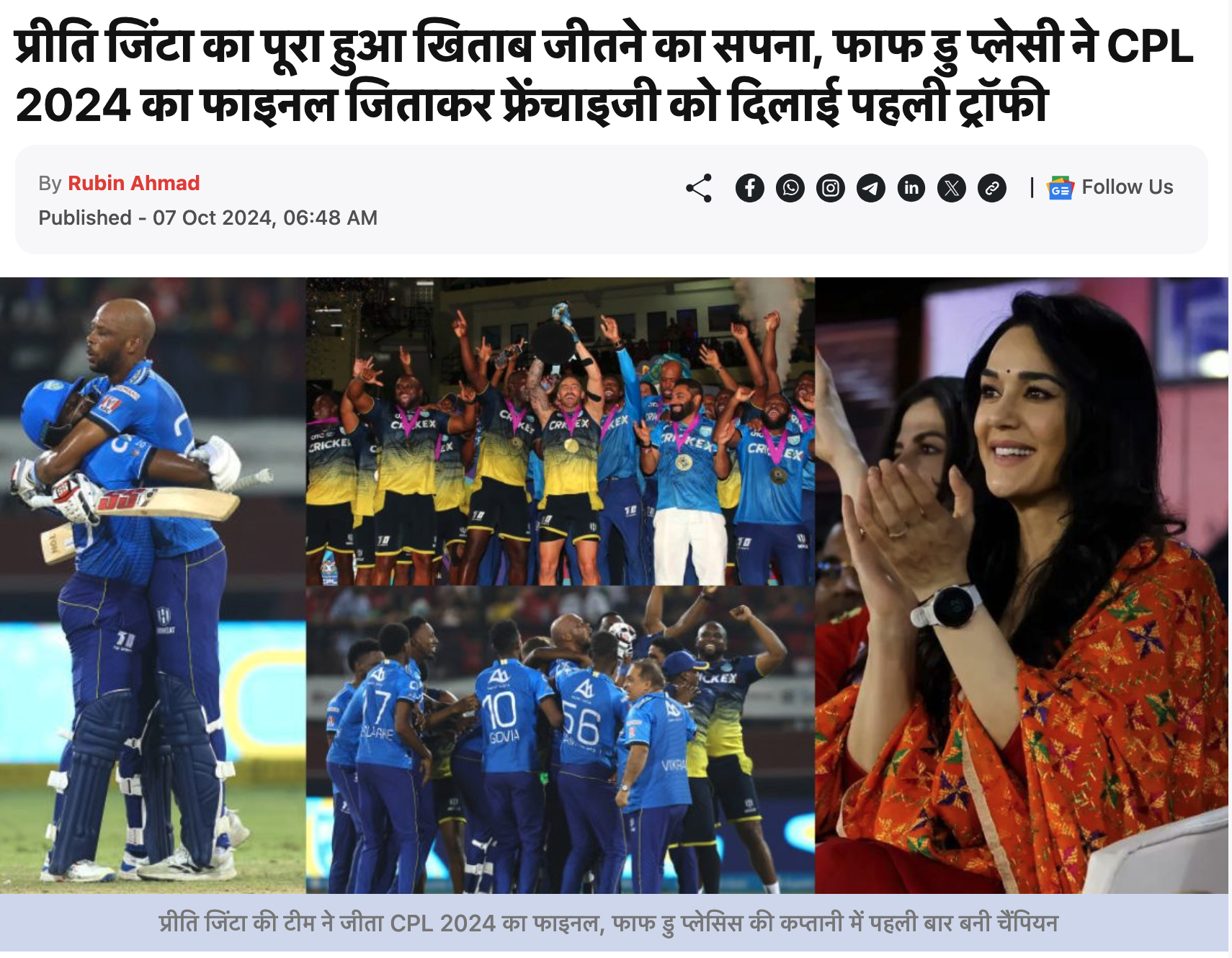} \\[0.5em]
\href{https://hindi.cricketaddictor.com/cricket-news/preity-zintas-team-saint-lucia-kings-won-the-1st-title-of-cpl-2024-under-the-captaincy-of-faf-du-plessis-7288602/}
{\parbox{7cm}{\small \textbf{Link:} https://hindi.cricketaddictor.com/cricket-news/preity-zintas-team-saint-lucia-kings-won-the-1st-title-of-cpl-2024-under-the-captaincy-of-faf-du-plessis-7288602/}} \\[0.5em]
{\parbox{7cm}{\small \textbf{Alt text:} \small ``Preity Zinta's team Saint Lucia Kings won the 1st title of CPL 2024 under the captaincy of Faf du Plessis''}}\\[0.5em]
\end{tabular}
&
\begin{tabular}{m{7cm}}
\centering
\includegraphics[width=7cm]{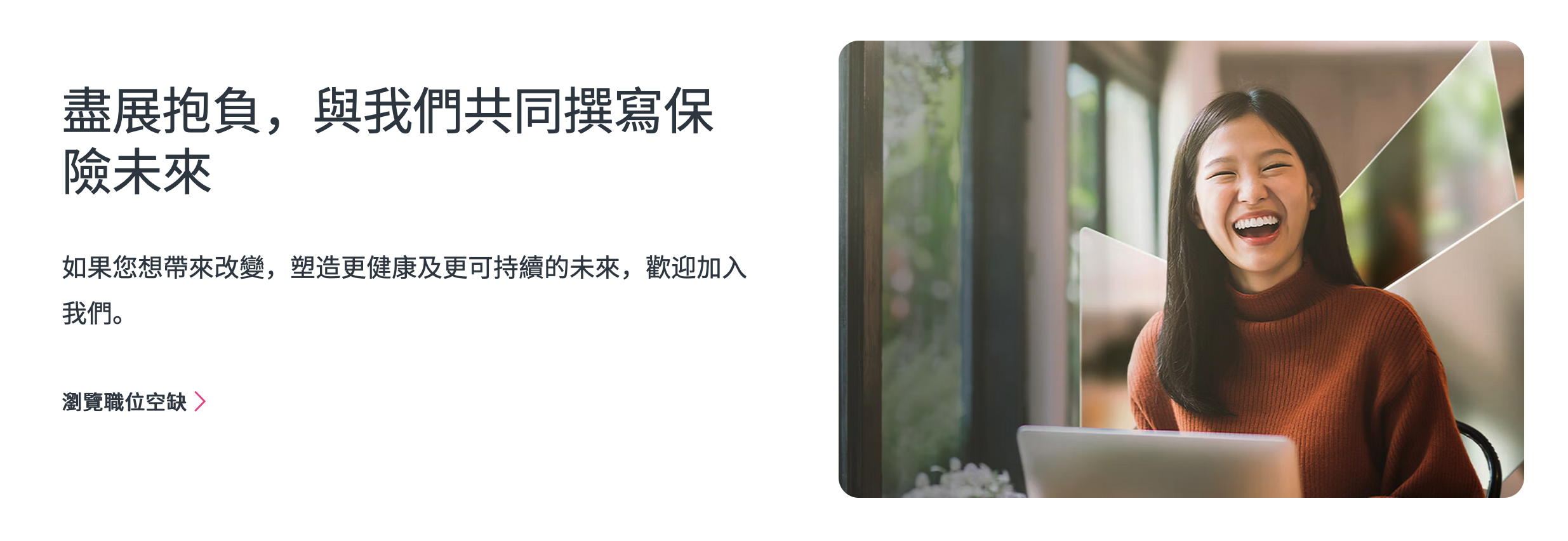} \\[0.5em]
\href{https://www.aia.com.hk}
{\parbox{7cm}{\small \textbf{Link:} https://www.aia.com.hk}} \\[0.5em]
{\parbox{7cm}{\small \textbf{Alt text:} \small ``Lovely Asian girl and mother celebrating birthday with friends / family by having a virtual birthday party at home. Enjoying birthday celebration in front of the laptop during pandemic. Practicing social distancing. Birthday lifestyle theme.''}}\\[0.5em]
\end{tabular}
\\
\hline

\end{tabular}
\caption{
Illustrative examples of accessibility mismatches on six websites across Bangladesh, India, Thailand, Egypt, China, and Hong Kong. Each cell consists of the image from the website, the corresponding URL, and the alt text or accessibility description associated with that image. The examples highlight cases where website content is presented in the native language, yet the accessibility descriptions, such as alt texts, are written in English.
}
\label{tab:mismatch_examples}
\end{table*}

\end{document}